%% file: main.tex
\documentclass{article}
\usepackage[T1]{fontenc}
\usepackage{newunicodechar}
\newunicodechar{–}{\textendash}
\newunicodechar{’}{\textquoteright}

\usepackage{iclr2027_conference,times}

\input{math_commands.tex}

\usepackage{amsmath}
\usepackage{amssymb}
\usepackage{booktabs}
\usepackage{tabularx}
\usepackage{graphicx}
\usepackage{multirow}
\usepackage{longtable}
\usepackage{algorithm}
\usepackage{algpseudocode}
\usepackage{xcolor}
\usepackage[skins,breakable]{tcolorbox}
\definecolor{PromptPurple}{HTML}{8B7BC4}
\definecolor{ExampleTeal}{HTML}{3B7F96}
\definecolor{ExampleTint}{HTML}{F0F9FC}
\newtcolorbox{paprompt}[1]{enhanced,breakable,colback=white,
  colframe=PromptPurple,colbacktitle=PromptPurple,coltitle=white,
  fonttitle=\fontfamily{lmss}\bfseries\small,
  fontupper=\fontfamily{lmss}\fontsize{9.4}{11.3}\selectfont,
  title={#1},title after break={#1\space(continued)},
  sharp corners,boxrule=0.9pt,left=6pt,right=6pt,top=5pt,bottom=5pt,
  toptitle=2pt,bottomtitle=2pt,before skip=8pt,after skip=10pt,
  before upper={\raggedright\setlength{\parindent}{0pt}\setlength{\parskip}{2.5pt}}}
\newtcolorbox{paexample}[1]{enhanced,breakable,colback=ExampleTint,
  colframe=ExampleTeal,colbacktitle=ExampleTeal,coltitle=white,
  fonttitle=\fontfamily{lmr}\bfseries\small,
  fontupper=\fontfamily{lmr}\small\selectfont,
  title={#1},title after break={#1\space(continued)},
  arc=2pt,boxrule=0.8pt,left=8pt,right=8pt,top=6pt,bottom=6pt,
  before skip=9pt,after skip=10pt,
  before upper={\setlength{\parindent}{0pt}\setlength{\parskip}{4pt}}}

\usepackage[hidelinks]{hyperref}
\usepackage{url}

\newcommand{\method}{\textsc{EncounterMob}}
\newif\iflogictrace
\logictracefalse

\iclrfinalcopy
\title{\resizebox{\textwidth}{!}{Encounter-Driven City-Scale Human Mobility}\\
Generation with Large Language Models}

\input{authors.tex}

\hypersetup{
  pdftitle={Encounter-Driven City-Scale Human Mobility Generation with Large Language Models},
  pdfauthor={Chao Li, Renhe Jiang, Chuan Xiao, Jiawei Wang}
}

\begin{document}

\maketitle
\lhead{Preprint. Under review.}

\input{sections/00_abstract}
\input{sections/01_introduction}

\input{sections/02_related_work}
\input{sections/03_problem_formulation}
\input{sections/04_metrics}
\input{sections/04_method}

\input{sections/05_experiments}
\input{sections/07_conclusion}

\newpage
\input{sections/08_statements}

\clearpage
\bibliography{references}
\bibliographystyle{iclr2027_conference}

\clearpage
\appendix
\input{sections/appendix}

\end{document}

%% file: math_commands.tex
\usepackage{amsmath,amsfonts,bm}

\def\eqref#1{equation~\ref{#1}}

\def\1{\bm{1}}

\DeclareMathAlphabet{\mathsfit}{\encodingdefault}{\sfdefault}{m}{sl}
\SetMathAlphabet{\mathsfit}{bold}{\encodingdefault}{\sfdefault}{bx}{n}



%% file: authors.tex
\author{%
Chao Li$^{1}$,\enspace Renhe Jiang$^{2}$,\enspace Chuan Xiao$^{3}$,\enspace Jiawei Wang$^{1}$\thanks{Corresponding author.}\\
\normalfont $^{1}$The Hong Kong University of Science and Technology\\
\normalfont $^{2}$The University of Tokyo,\enspace $^{3}$The University of Osaka\\
\normalfont \texttt{chaoli@hkust-gz.edu.cn},\enspace \texttt{jiangrh@csis.u-tokyo.ac.jp}\\
\normalfont \texttt{chuanx@ist.osaka-u.ac.jp},\enspace \texttt{jiaweiw@hkust-gz.edu.cn}
}

%% file: sections/00_abstract.tex
\begin{abstract}
    Large language models (LLMs) have shown strong potential for human mobility generation by modeling individual routines, activities, and mobility preferences. However, existing approaches remain largely resident-centric: generating realistic individual trajectories does not necessarily recover the encounters that emerge when multiple residents co-occur at the same place and time. To this end, we study encounter-aware human mobility generation, which jointly preserves individual mobility and population-level interaction structure. We propose \method{}, an encounter-driven LLM framework consisting of a coarse-to-fine Encounter Generator and an Iterative Coupling Framework. The Encounter Generator first captures city-level encounter rhythms and then instantiates participant--time--venue encounters, while iterative coordinates these encounters with personal mobility behavior through successive refinement. We further introduce an encounter recovery protocol covering encounter marginal and joint distributions over residents, times, and venues. Experiments on the Japan and NYC datasets show that \method{} outperforms all external baselines on all seven encounter metrics and all six personal mobility metrics. Compared with ELLMob, it reduces the participant--time--venue joint error by 40.4\%/41.7\% and the mean personal mobility error by 58.8\%/32.7\% on Japan/NYC, respectively.
\end{abstract}

%% file: sections/01_introduction.tex
\section{Introduction}

City-scale human mobility data capture the spatiotemporal movements of residents across cities and support the study of individual mobility patterns and urban dynamics, with established applications in transportation planning \citep{colak2016}, public-health decision-making \citep{chang2021}, and urban planning and management \citep{zheng2014}. However, access to real-world mobility data are often constrained by collection costs, limited coverage, commercial restrictions, and privacy risks \citep{kostandova2025,synmob2023}. Human mobility generation therefore aims to synthesize spatiotemporal trajectories that reproduce real-world mobility patterns, alleviating data scarcity and supporting downstream mobility analysis and urban applications\citep{worldmove2026}.

Recent LLM-based mobility generation methods have substantially improved individual trajectory synthesis by leveraging rich behavioral information, including historical routines\citep{llmob2024}, activity semantics\citep{mobilityllm2024}, mobility motivations\citep{trajllm2025}, and contextual factors \citep{ellmob2026}. Beyond reproducing spatiotemporal regularities, these methods use LLMs to interpret the behavioral semantics underlying individual mobility, such as what activities residents perform, what places they prefer, and how their mobility adapts to different contexts. Such semantic reasoning enables generated trajectories to better reflect not only where and when each resident moves, but also the behavioral patterns, preferences, and motivations that shape their movement.

\begin{figure*}[!t]
    \centering    
    \includegraphics[width=\textwidth]{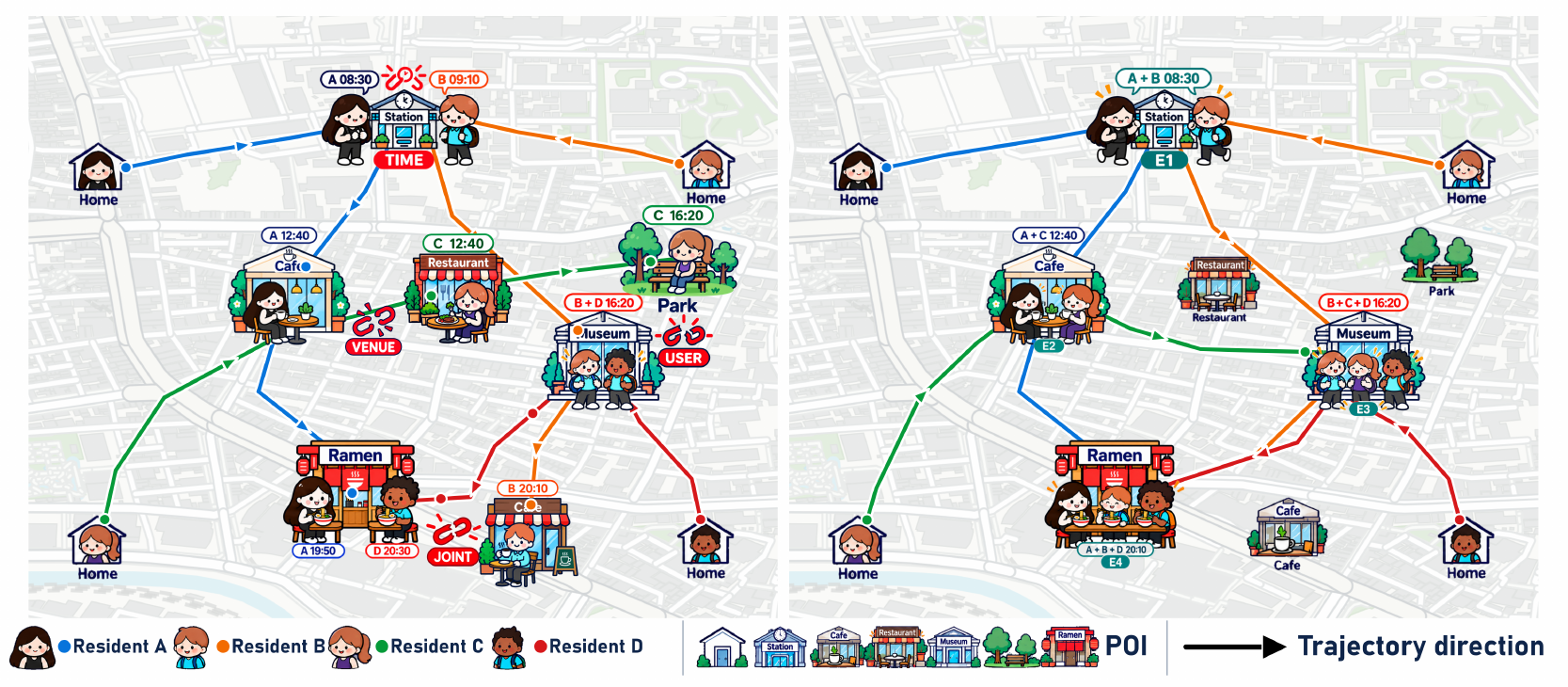}
    \caption{Individual mobility generation versus encounter-driven mobility generation. Colored paths represent the trajectories of different residents, and highlighted locations indicate encounters across trajectories. (a) Individual mobility generation models each resident according to personal mobility patterns. (b) Encounter-driven mobility generation additionally coordinates encounter participants, times, and locations across the trajectories of multiple residents within the generated mobility data.}
    \label{fig:encounter_thesis}
\end{figure*}

Despite their remarkable success, existing methods primarily adopt a resident-centric perspective, generating individual trajectories according to personal mobility behavioral. However, mobility data also capture cross-resident relations, including encounters, i.e., the co-presence of two or more residents at the same place and time \citep{dailyencounters2013}. Preserving such encounters requires coordination across multiple residents, since the activity of one resident must align with the timing, location, and activity context of others. Recent household mobility work uses LLMs to reason about activity dependencies among household members and coordinate their mobility accordingly \citep{householdllm2025}. However, in existing city-scale mobility generation, encounters are typically induced by spatiotemporal overlaps among independently generated trajectories. Since encounters reflect the joint behavior of multiple residents, resident-centric generation may be insufficient to capture the collective interactions involved in such events \citep{deepmobility2025}. As a result, the ability of existing methods to recover encounter structures remains questionable. As illustrated in Figure~\ref{fig:encounter_thesis}(a), independently generated trajectories may produce mismatches in encounter participants, times, and venues, leading to inaccurate city-wide encounter distributions across these dimensions.

Therefore, we propose \method{}, an encounter-driven approach to human mobility generation that coordinates resident trajectories to reproduce encounter structures in the generated mobility data, as illustrated in Figure~\ref{fig:encounter_thesis}(b). This formulation introduces two key challenges. First, the encounter structures used for coordination need to be established before generating individual trajectories. We introduce a coarse-to-fine encounter generator that first captures the temporal and category-level distribution of encounters and then instantiates concrete participant--time--venue events. Second, these encounter structures need to be coupled with the mobility patterns of individual residents. We therefore introduce an iterative coupling framework, where a joint critic evaluates their consistency and guides the refinement of both encounter structures and personal trajectories. Together, these components enable coordinated trajectory generation while preserving plausible personal behavior. Besides, we introduce an encounter-aware evaluation protocol that assesses encounter recovery across participants, times, and venues, together with their joint structures.

Experiments on the Japan and NYC datasets demonstrate consistent improvements in both encounter and personal mobility fidelity. Compared with ELLMob, \method{} achieves lower scores on all seven encounter metrics, reducing the joint encounter score by 40.4\%/41.7\%(Japan/NYC), and on all six personal mobility metrics, with mean reductions of 58.8\%/32.7\%, respectively. Ablation studies show that removing either city-level rhythms or Encounter Instantiation increases the full group--time--venue JSD by 33.6\% on average across the two ablations and both datasets. Removing iterative coupling increases this score by 74.4\% on average across the two datasets, supporting the contributions of encounter generation and iterative refinement to encounter recovery. Our contributions are summarized as follows:

\begin{enumerate}
    \item To the best of our knowledge, we present the first city-scale human mobility generation study aimed at recovering the encounter structures observed in real mobility data.
    
    \item We propose \method{}, an encounter-driven LLM framework that generates encounter structures and coordinates them with personal mobility behavior.
    
    \item We develop an encounter-aware evaluation protocol that measures participant, temporal, and spatial structures of generated encounters and their dependencies.
    
    \item Extensive experiments on Japan and NYC show that \method{} outperforms all external baselines across all encounter and personal mobility metrics.
\end{enumerate}

%% file: sections/02_related_work.tex
\section{Related Work}
\paragraph{City-Scale Human Mobility Generation}
Deep learning has been widely studied for human mobility prediction and generation \citep{luca2023}. Early neural methods use recurrent architectures to capture temporal dependencies in trajectories \citep{kulkarni2017rnn,berke2022rnn}, while later studies employ variational models, GANs, adversarial imitation learning, and diffusion models for trajectory generation \citep{svae2019,movesim2020,trajgail2021,difftraj2023,codiffmob2025}. More recently, LLM-based methods incorporate behavioral and contextual information into individual trajectory generation, including activity patterns and motivations \citep{llmob2024}, persona-conditioned activity planning \citep{trajllm2025}, and event-specific context \citep{ellmob2026}. These advances establish LLMs as a promising direction for human mobility generation.

\paragraph{LLMs for Human Mobility Modeling}
LLM-based approaches to human mobility use mobility histories and behavioral context to reason about individual movement \citep{llmmob2023,mobilityllm2024}. Recent frameworks further incorporate demographics, activities, motivations, preferences, and external events to support interpretable trajectory generation and behavioral adaptation \citep{mobglm2024,llmob2024,trajllm2025,ellmob2026}. Beyond individual mobility, household mobility work uses LLMs to infer activity dependencies among household members and coordinate their schedules \citep{householdllm2025}. These studies extend LLM-based mobility modeling from individual behavior to coordinated and population-level mobility across multiple scales. However, existing city-scale mobility generation primarily treats resident independently, with encounter structures arising only from spatiotemporal overlap rather than cross-resident coordination.

%% file: sections/03_problem_formulation.tex
\section{Problem Formulation}
\label{sec:problem}

Given historical mobility records and the set of active residents $\mathcal{R}_{\mathrm{act}}$ for the target day, we aim to generate a daily trajectory $\widehat{\tau}_r$ for each resident $r\in\mathcal{R}_{\mathrm{act}}$. Each trajectory records a time-ordered sequence of venue visits. When these trajectories place two or more residents at the same venue during the same time interval, they form an encounter, denoted by $e=(t_e,v_e,\mathcal{R}_e)$, where $t_e$, $v_e$, and $\mathcal{R}_e$ specify the time, venue, and complete set of co-present residents, respectively. For a set of trajectories $\mathcal{T}$, let $\mathcal{E}(\mathcal{T})$ denote all encounters induced by $\mathcal{T}$. Their organization across participants, times, and venues defines the encounter structure of $\mathcal{T}$. Formally, our task is to generate the joint trajectory set $\widehat{\mathcal{T}}=\{\widehat{\tau}_r\mid r\in\mathcal{R}_{\mathrm{act}}\}$ such that each trajectory preserves the mobility regularities of its resident, while $\mathcal{E}(\widehat{\mathcal{T}})$ reproduces the encounter structure observed in real mobility data.

%% file: sections/04_metrics.tex
\section{Evaluation Protocol for Encounter Recovery}
\label{sec:metrics}
Existing evaluation of human mobility generation mainly assesses mobility
patterns at the individual or population level
\citep{llmob2024,ellmob2026,codiffmob2025,deepmobility2025}. However, these
two levels capture different aspects of mobility fidelity and do not evaluate
the cross-resident structure formed by encounters. For example, generated
trajectories may reproduce individual mobility patterns and aggregate
time--venue distributions, while different groups of residents meet at those
times and venues. To this end, we propose a evaluation protocol
to evaluate encounter recovery across participants, times, and
venues, including their marginal and joint distributions.

\begin{table}[h]
\centering
\caption{Evaluation dimensions in the proposed encounter recovery protocol.}
\label{tab:encounter_dimensions}
\small
\setlength{\tabcolsep}{4pt}
\begin{tabular}{@{}llp{0.48\linewidth}@{}}
\toprule
Dimension & Distribution & Characterization \\
\midrule
Participant group
& $G$ (GroupD)
& Complete participant groups involved in encounters \\

Time
& $T$ (TD)
& Temporal distribution of encounters \\

Venue
& $V$ (VD)
& Venue distribution of encounters \\

Participant--Time
& $G{\times}T$
& Association between participant groups and times \\

Participant--Venue
& $G{\times}V$
& Association between participant groups and venues \\

Time--Venue
& $T{\times}V$
& Association between encounter times and venues \\

Full joint
& $G{\times}T{\times}V$
& Joint participant--time--venue structure \\
\bottomrule
\end{tabular}
\end{table}

Specifically, each encounter is jointly characterized by its complete
participant group $G$, time $T$, and venue $V$. We evaluate encounter recovery
using seven distributions, as summarized in Table~\ref{tab:encounter_dimensions}.
The marginal distributions of $G$, $T$, and $V$ assess which residents
participate in encounters, when encounters occur, and where they occur. The
joint distributions $G{\times}T$, $G{\times}V$, and $T{\times}V$ evaluate the
associations between these dimensions, while $G{\times}T{\times}V$ evaluates
their participant--time--venue structure. Together, these dimensions provide a comprehensive evaluation of encounter recovery.

Following prior mobility-generation evaluation
\citep{llmob2024,ellmob2026}, after extracting the above distributions from
both generated and ground-truth mobility data, Jensen--Shannon divergence
(JSD) is employed to quantify the discrepancy between them. Details are provided in
Appendix~\ref{app:metric_details}.

%% file: sections/04_method.tex
\section{Methodology}
\label{sec:method}

\begin{figure*}[t]
    \centering
    \includegraphics[width=\textwidth]{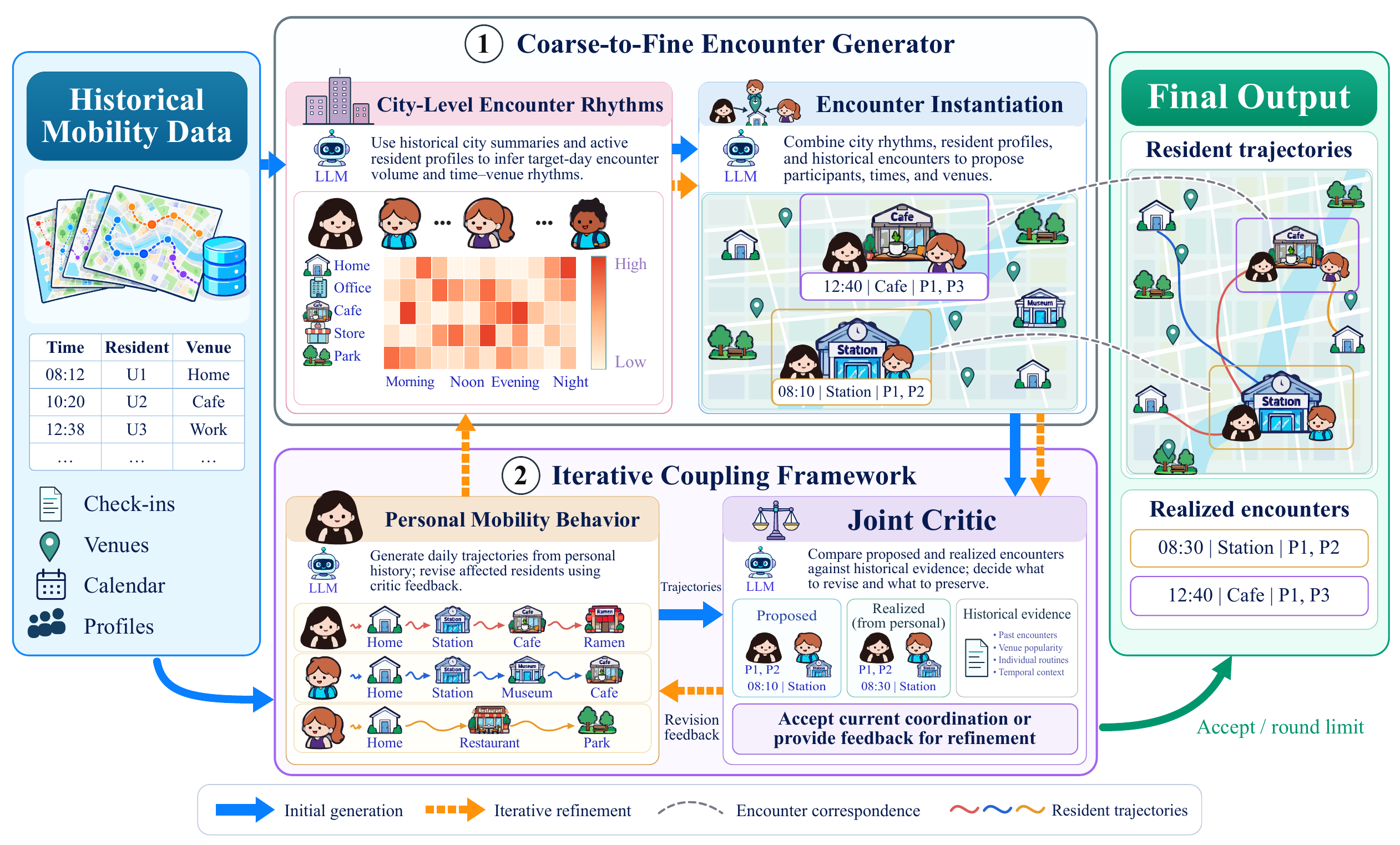}
    \caption{Overview of \method{}. (1) The coarse-to-fine Encounter Generator infers city-level encounter rhythms from historical mobility and resident profiles, then instantiates participant--time--venue encounter proposals. (2) The Iterative Coupling Framework compares these proposals with encounters realized by independently initialized personal trajectories. Guided by historical evidence, the joint critic provides feedback to revise personal trajectories and, when needed, city-level rhythms and encounter proposals. The personal trajectories form the final output. Colored dashed curves illustrate correspondences between proposed and realized encounters after coupling.}

    \label{fig:framework}
\end{figure*}

Figure~\ref{fig:framework} presents \method{}, an encounter-driven LLM framework for city-scale human mobility generation. It consists of a coarse-to-fine Encounter Generator and an Iterative Coupling Framework. The Encounter Generator derives concrete participant--time--venue encounters from city-level encounter rhythms, while the Iterative Coupling Framework iteratively adjusts encounter proposals and personal mobility behaviors. Together, the two components guide resident trajectories toward realistic encounter structures while preserving individual mobility regularities.

\subsection{Coarse-to-Fine Encounter Generator}

To guide mobility generation with encounters, we need to identify plausible encounters for the active residents. We introduce a coarse-to-fine Encounter Generator that first infers city-level encounter rhythms and then instantiates them into concrete participant--time--venue encounter proposals.

\paragraph{City-Level Encounter Rhythms}

Encounters arise as residents carry out their daily activities across the city. Since these activities follow recurring temporal and spatial routines, the resulting encounters also exhibit regularities in when and where they occur~\citep{leng2021collective}. For example, commuting brings many residents to stations during the morning and evening rush hours, while meal periods concentrate residents in restaurants, creating more opportunities for encounters at these times and places. Based on this, we characterize the overall encounter activity of the target day as city-level encounter rhythms, providing coarse guidance on the times and venue categories in which encounters are likely to occur.

To infer these rhythms for a specific target day, we consider both historical encounter activity and the behavioral context of the active residents. Historical encounter activity is summarized from several aspects, including daily encounter volume, temporal distribution, venue-category composition, and weekday variation. For each active resident, we further summarize historical mobility into a behavioral profile describing recurring activities together with their typical timing and venue conditions. The LLM combines these sources to determine how the historical encounter regularities should be reflected for the current population, producing the target-day encounter volume and its coarse distribution over time and venue categories.

\paragraph{Encounter Instantiation}

Given the city-level encounter rhythms, the next step is to instantiate specific encounters for the target day. In addition to the resident profiles used above, we summarize historical encounters through their participants, meeting times, venues, and activity contexts, providing more specific evidence for how residents have interacted in the past. The LLM combines this historical evidence with the coarse temporal and venue-category guidance to determine plausible participant--time--venue configurations. Historical relations support previously observed combinations, while the behavioral context of the residents also allows the LLM to consider new combinations when they are consistent with their activities. The resulting instantiated encounters form the initial encounter proposals for the subsequent Iterative Coupling Framework.

\subsection{Iterative Coupling Framework}

The initial encounter proposals need to be further coordinated with the mobility behavior of the involved residents on the target day. A proposed encounter may require a resident to change an otherwise plausible activity, while established personal mobility patterns may make the proposed time, venue, or participant combination difficult to realize. We therefore treat both encounter proposals and personal mobility behavior as revisable. Then an Iterative Coupling Framework is introduced to enable them to couple each other through iterative feedback~\citep{selfrefine2023,gou2024critic}.

First, we prompt the LLM to generate a complete daily trajectory for each resident based on historical personal mobility patterns, and then extract the encounters formed by these trajectories. Second, the LLM compares these extracted encounters with the encounter proposals and examines the historical evidence supporting both sides. For personal behavior, such evidence includes recurring activities and their typical timing and venue conditions; for encounter proposals, it includes the historical encounter contexts and resident activities used during instantiation. Based on this comparison, the Joint Critic either accepts the current coordination or identifies the remaining inconsistencies and provides feedback for refinement.

At each refinement round, the Critic feedback is passed to the personal mobility generator to revise the affected trajectories while preserving personal behavior consistent with the supported encounter proposals. The revised trajectories are then fed back to the coarse-to-fine Encounter Generator, which updates the city-level encounter rhythms and subsequently regenerates the encounter proposals conditioned on the updated rhythms and current personal trajectories. The revised trajectories and encounter proposals are evaluated again by the Joint Critic in the next round. Through this process, personal behavior can adapt to supported encounter proposals, while encounter proposals can also be revised according to well-supported personal behavior. The refinement proceeds for at most \(K\) rounds and terminates earlier once the Joint Critic accepts the current coordination.

%% file: sections/05_experiments.tex
\section{Experiments}

\subsection{Experimental Setup}

\paragraph{Datasets.}
We use two city-scale mobility datasets constructed from the global Foursquare check-in resource associated with LBSN2Vec \citep{yang2019lbsn,lbsndata2021}, covering Japan and the New York metropolitan area. The Japan and NYC datasets use fixed cohorts of 1,049 and 226 residents constructed solely from historical-period encounters, with 60/30 and 123/61 historical/test days, respectively.  Both datasets are discretized into 10-minute intervals\citep{llmob2024,ellmob2026} and use Foursquare venue categories. More detailed information is provided in Appendix~\ref{app:dataset_details}.

\paragraph{Metrics.}
We evaluate both encounter and personal mobility fidelity. For encounters, we use the metrics defined in Section~\ref{sec:metrics}: TD, VD, GroupD, $T{\times}V$, $G{\times}T$, $G{\times}V$, and $G{\times}T{\times}V$. For personal mobility, we use six metrics from LLMob and ELLMob \citep{llmob2024,ellmob2026}: step distance (SD) and step interval (SI) measure consecutive-stop distances and time gaps; daily activity routine distribution (DARD) and spatio-temporal visit distribution (STVD) capture time--activity and time--location patterns; category distribution (CD) and spatial grid distribution (SGD) capture activity composition and spatial visit distributions. All scores compare normalized observed and generated histograms using Jensen--Shannon divergence (JSD) \citep{lin1991jsd}; lower is better. Detailed definitions and calculation settings are provided in Appendix~\ref{app:metric_details}.

\paragraph{Baselines.}
We compare \method{} with two types of baselines: \textit{(1) deep-learning-based methods}, including predictive models LSTM~\citep{lstm1997}, GETNext~\citep{getnext2022}, and MHSA~\citep{mhsa2023}, and generative models TrajGAIL~\citep{trajgail2021} and DiffTraj~\citep{difftraj2023}; and \textit{(2) LLM-based methods}, including LLMob~\citep{llmob2024} and ELLMob~\citep{ellmob2026}. Within each dataset, all methods use the same settings, information, and evaluation metrics; the LLM-based methods use the same backbone unless otherwise specified. Implementation and generation details are provided in Appendix~\ref{app:implementation_details}.

\subsection{Main Results and Analysis}

\begin{table*}[t]
\centering
\caption{Encounter fidelity on Japan and NYC. All metrics are JSDs computed with natural logarithms on normalized histograms; lower is better. Bold and underline indicate the best and second-best results among the compared methods; ablations are excluded from ranking.}
\label{tab:encounter_main}
\label{tab:encounter_nyc}
\small
\setlength{\tabcolsep}{3pt}
\renewcommand{\arraystretch}{1.10}
\resizebox{\textwidth}{!}{
\begin{tabular}{@{}lrrrrrrr@{\hspace{8pt}}rrrrrrr@{}}
\toprule
& \multicolumn{7}{c}{Japan} & \multicolumn{7}{c}{NYC} \\
\cmidrule(lr){2-8}\cmidrule(lr){9-15}
Model & TD & VD & GroupD & $T{\times}V$ & $G{\times}T$ & $G{\times}V$ & $G{\times}T{\times}V$ & TD & VD & GroupD & $T{\times}V$ & $G{\times}T$ & $G{\times}V$ & $G{\times}T{\times}V$ \\
\midrule
LSTM & 0.09877 & 0.24476 & 0.68514 & 0.50963 & 0.69207 & 0.68664 & 0.69207 & 0.69315 & 0.66362 & 0.69315 & 0.69315 & 0.69315 & 0.69315 & 0.69315 \\
GETNext & \underline{0.05961} & 0.23111 & 0.68071 & 0.49085 & 0.69315 & 0.68496 & 0.69315 & 0.65010 & 0.69315 & 0.69315 & 0.69315 & 0.69315 & 0.69315 & 0.69315 \\
MHSA & 0.06601 & 0.22904 & 0.68160 & 0.49340 & 0.69315 & 0.68468 & 0.69315 & 0.64424 & 0.66869 & 0.69315 & 0.69315 & 0.69315 & 0.69315 & 0.69315 \\
TrajGAIL & 0.08504 & 0.45497 & 0.68895 & 0.60298 & 0.69315 & 0.69167 & 0.69315 & 0.45976 & 0.66984 & 0.66637 & 0.69315 & 0.69315 & 0.69315 & 0.69315 \\
DiffTraj & 0.06280 & 0.22032 & 0.67645 & 0.46585 & 0.69286 & 0.68045 & 0.69286 & 0.66286 & 0.66728 & 0.69315 & 0.69315 & 0.69315 & 0.69315 & 0.69315 \\
LLMob & 0.22617 & 0.23132 & 0.65848 & 0.51287 & 0.68804 & 0.66946 & 0.68804 & 0.58444 & 0.69315 & 0.54434 & 0.69315 & 0.66869 & 0.69315 & 0.69315 \\
ELLMob & 0.07869 & \underline{0.19624} & \underline{0.56109} & \underline{0.36724} & \underline{0.59846} & \underline{0.57975} & \underline{0.59938} & \underline{0.45658} & \underline{0.54306} & \underline{0.43226} & \underline{0.55025} & \underline{0.54232} & \underline{0.54306} & \underline{0.55025} \\
\method{} & \textbf{0.03653} & \textbf{0.10779} & \textbf{0.30178} & \textbf{0.22091} & \textbf{0.35382} & \textbf{0.33267} & \textbf{0.35742} & \textbf{0.20716} & \textbf{0.29093} & \textbf{0.14600} & \textbf{0.32079} & \textbf{0.27262} & \textbf{0.29093} & \textbf{0.32079} \\
\midrule
w/o Rhythm & 0.03844 & 0.12302 & 0.35228 & 0.23459 & 0.39645 & 0.38434 & 0.39952 & 0.17373 & 0.32835 & 0.13937 & 0.38107 & 0.28658 & 0.32835 & 0.38107 \\
w/o Instantiation & 0.04170 & 0.19092 & 0.45823 & 0.34306 & 0.52240 & 0.49942 & 0.52344 & 0.25463 & 0.44168 & 0.23567 & 0.47931 & 0.47926 & 0.48400 & 0.50454 \\
w/o Coupling & 0.06393 & 0.16443 & 0.52829 & 0.33275 & 0.56156 & 0.54584 & 0.56274 & 0.53722 & 0.58943 & 0.47331 & 0.61383 & 0.59937 & 0.60546 & 0.61383 \\
\bottomrule
\end{tabular}}
\end{table*}

\begin{table*}[t]
\centering
\caption{Personal mobility fidelity on Japan and NYC. All metrics are JSDs computed with natural logarithms; lower is better. Bold and underline indicate the best and second-best results among the compared methods; ablations are excluded from ranking.}
\label{tab:personal_main}
\label{tab:personal_nyc}
\small
\setlength{\tabcolsep}{3pt}
\renewcommand{\arraystretch}{1.10}
\resizebox{\textwidth}{!}{
\begin{tabular}{@{}lrrrrrr@{\hspace{8pt}}rrrrrr@{}}
\toprule
& \multicolumn{6}{c}{Japan} & \multicolumn{6}{c}{NYC} \\
\cmidrule(lr){2-7}\cmidrule(lr){8-13}
Model & SD & SI & DARD & STVD & CD & SGD & SD & SI & DARD & STVD & CD & SGD \\
\midrule
LSTM & 0.06371 & 0.10285 & 0.17937 & 0.58511 & 0.00493 & 0.11069 & 0.22653 & 0.11680 & 0.58435 & 0.68255 & 0.00870 & 0.20886 \\
GETNext & 0.05912 & 0.08853 & 0.15050 & 0.55073 & 0.01972 & 0.10460 & 0.25078 & 0.16924 & 0.57884 & 0.68183 & \underline{0.00541} & 0.20827 \\
MHSA & 0.06143 & 0.05722 & 0.16247 & 0.55529 & 0.01929 & 0.11078 & 0.25595 & 0.15803 & 0.55372 & 0.67848 & 0.00734 & 0.19664 \\
TrajGAIL & 0.12680 & 0.08676 & 0.22410 & 0.60217 & 0.03489 & 0.22519 & 0.37290 & 0.20186 & 0.60749 & 0.68741 & 0.03541 & 0.35073 \\
DiffTraj & 0.08385 & 0.09709 & 0.14790 & 0.53861 & 0.01787 & 0.10721 & 0.23151 & 0.17015 & 0.55110 & 0.67917 & 0.01012 & 0.20594 \\
LLMob & 0.07955 & 0.14059 & 0.22595 & 0.58424 & 0.00638 & 0.12113 & 0.19057 & 0.15416 & 0.53990 & 0.67073 & 0.01240 & 0.21916 \\
ELLMob & \underline{0.00931} & \underline{0.01112} & \underline{0.09474} & \underline{0.35217} & \underline{0.00121} & \underline{0.05058} & \underline{0.07621} & \underline{0.05242} & \underline{0.30089} & \underline{0.39345} & 0.00583 & \underline{0.07278} \\
\method{} & \textbf{0.00429} & \textbf{0.00331} & \textbf{0.03391} & \textbf{0.15189} & \textbf{0.00037} & \textbf{0.01989} & \textbf{0.05221} & \textbf{0.02475} & \textbf{0.20448} & \textbf{0.27555} & \textbf{0.00117} & \textbf{0.04836} \\
\midrule
w/o Rhythm & 0.00495 & 0.00462 & 0.04027 & 0.17562 & 0.00044 & 0.02420 & 0.05561 & 0.02735 & 0.21856 & 0.29499 & 0.00202 & 0.05280 \\
w/o Instantiation & 0.00566 & 0.00456 & 0.03828 & 0.17327 & 0.00039 & 0.02342 & 0.05883 & 0.02906 & 0.22089 & 0.30294 & 0.00133 & 0.05447 \\
w/o Coupling & 0.01005 & 0.01146 & 0.07809 & 0.30375 & 0.00172 & 0.05049 & 0.08013 & 0.04613 & 0.31159 & 0.41470 & 0.00331 & 0.08346 \\
\bottomrule
\end{tabular}}
\end{table*}

\paragraph{Main results.}
Tables~\ref{tab:encounter_main} and~\ref{tab:personal_main} evaluate encounter and personal mobility fidelity on Japan and NYC. \method{} outperforms all external baselines on all seven encounter metrics and all six personal metrics. Several baseline scores reach the maximum JSD value of \(0.69315\), indicating no overlap between the observed and generated distributions. ELLMob is competitive on personal mobility, leading all six baseline metrics in Japan and five in NYC, yet its encounter errors remain substantially higher than those of \method{}. This shows that personal fidelity alone is not enough to assess generated mobility, and encounter recovery provides a complementary view.

The encounter improvements extend from temporal and venue distributions to complete participant groups and their joint configurations. Compared to ELLMob, \method{} reduces GroupD by 46.2\% in Japan and 66.2\% in NYC, and reduces $G{\times}T{\times}V$ by 40.4\% and 41.7\%, respectively. These results support encounter modeling for recovering which residents meet, when, and where.

These encounter gains accompany improvements in personal mobility. Compared with ELLMob, \method{} reduces the mean of the six personal errors by 58.8\% in Japan and 32.7\% in NYC, with lower errors on every personal metric in both datasets. The complete framework therefore improves encounter recovery while also preserving more faithfully individual mobility, suggesting that correctly formed encounters can in turn benefit the quality of individual mobiltiy generation.

\paragraph{Ablation study.}
We conduct ablation studies to examine three key designs in \method{}: city-level encounter rhythms (w/o Rhythm), Encounter Instantiation (w/o Instantiation), and the Iterative Coupling Framework (w/o Coupling). As can be seen in Table 2, removing city-level encounter rhythms consistently degrades encounter generation performance on both datasets. The improvement on TD, VD, and \(T\times V\) is the most significant, showcasing the importance of city-level rhythms in recovering the temporal and venue distributions of encounters. We also observe improvements in GroupD on both datasets. This effectiveness is likely attributed to more accurate temporal and venue guidance, which reduces inconsistency in the resulting group distribution.

The importance of Encounter Instantiation is particularly evident in GroupD and the group-related joint metrics, where its removal causes consistent performance degradation on both datasets. This highlights its critical role in determining which residents participate in encounters and how they are associated with time and venue. These results also indicate that city-level encounter rhythms alone are insufficient to recover concrete participant–time–venue configurations.

By comparing \method{} with w/o Coupling, we observe clear improvements in GroupD and the group-related joint metrics on both datasets, highlighting the role of iterative coupling in coordinating encounter requirements with personal mobility behavior. Meanwhile, the mean of the six personal mobility errors decreases by 53.1\% in Japan and 35.4\% in NYC, showing that improved encounter fidelity is accompanied by better individual mobility fidelity.

\subsection{Recovery of City-Level Encounter Rhythms Distributions}

\begin{figure*}[t]
\centering
\includegraphics[width=\textwidth]{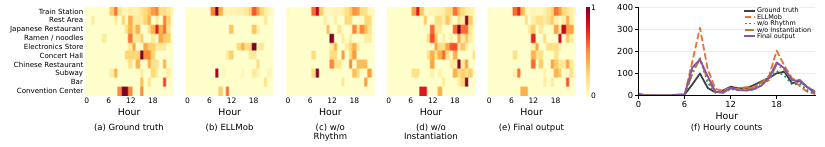}
\caption{City-level encounter patterns on Japan over 30 days. (a--e) Hourly encounter distributions across the ten most frequent ground-truth venue categories for ground truth, ELLMob, w/o Rhythm, w/o Instantiation, and final output. Each heatmap is normalized within these categories; colors are scaled by the maximum for each category across all five panels and hours. (f) Hourly encounter counts across all venue categories for the same five sources, pooled over the 30 days. Generated encounters are extracted from final trajectories.}
\vspace{-4mm}
\label{fig:city_pattern}
\end{figure*}

Figure~\ref{fig:city_pattern} compares the encounter distributions in ground truth, ELLMob, our city-level encounter rhythms, and the final generated trajectories. The heatmaps show that ELLMob deviates from ground truth in how encounters are distributed across hours and venue categories, whereas our city-level encounter rhythms more closely capture the observed distribution. The hourly counts in Figure~\ref{fig:city_pattern}(f) reveal a similar difference in encounter volume: ELLMob produces an overly concentrated peak, while our city-level encounter rhythms better follow the temporal variation in ground truth. The encounters extracted from the final trajectories largely preserve this distribution, while the adjustments introduced during coupling generally bring both the category allocation and hourly counts closer to ground truth. These results show that city-level encounter rhythms provide effective guidance for recovering the encounter distribution while allowing adaptation to personal mobility.

\begin{table}[h]
\centering
\caption{Participant-group recovery in Japan and NYC. Count is the number of generated encounter events, precision, recall, and F1 are percentages; higher is better.}
\label{tab:participant_identity}
\small
\setlength{\tabcolsep}{5pt}
\begin{tabular}{llrrrr}
\toprule
Dataset & Method & Count
& Precision $\uparrow$
& Recall $\uparrow$
& F1 $\uparrow$ \\
\midrule
Japan & ELLMob & 1,627 & 13.83 & 15.53 & 14.63 \\
 & w/o Rhythm & 940 & 44.55 & 38.75 & 41.45 \\
 & w/o Instantiation & 1,104 & 27.86 & 28.73 & 28.29 \\
 & \method{} & 1,111 & \textbf{49.66} & \textbf{49.86} & \textbf{49.76} \\
\midrule
NYC & ELLMob & 26 & 50.00 & 30.77 & 38.10 \\
 & w/o Rhythm & 123 & 69.23 & \textbf{69.23} & \textbf{69.23} \\
 & w/o Instantiation & 88 & 31.82 & 53.85 & 40.00 \\
 & \method{} & 80 & \textbf{71.43} & 57.69 & 63.83 \\
\bottomrule
\end{tabular}
\end{table}

\subsection{Recovery of Participant Identities}
\label{sec:participant_identity}
Beyond predicting when and where encounters occur, Encounter Instantiation must also identify the full set of residents involved in each encounter. The group-based distributional metrics used in the main evaluation evaluate the distributions of participant groups, but they do not directly measure whether individual groups are recovered exactly. We therefore also report group precision, recall, and F1 under exact matching. A predicted group is counted as correct only when its full participant set exactly matches a group in the ground-truth data. As shown in Table~\ref{tab:participant_identity}, Encounter Instantiation improves participant-group recovery across both datasets. These results highlight its role in recovering the complete sets of residents involved in encounters. In NYC, w/o Rhythm achieves higher Recall and F1, but only by generating more encounters.

\subsection{Effect of Iterative Coupling}
\label{sec:iterative_coupling}
\begin{figure*}[h]
\centering
\includegraphics[width=0.65\textwidth]{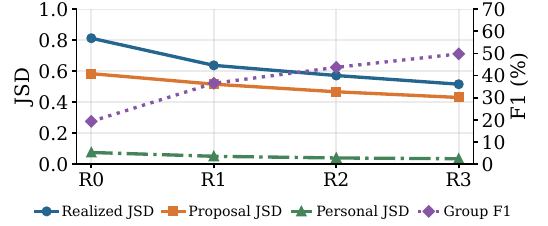}
\caption{Effect of Iterative Coupling on Japan over R0--R3. Encounter JSD decreases across rounds, while group recovery F1 of the generated trajectories increases and personal mean JSD remains relatively stable. JSD curves use the left axis; group recovery F1 uses the right axis. Lower JSD and higher F1 are better.}
\label{fig:coupling_roundwise}
\end{figure*}

We investigate the effect of refinement rounds in the Iterative Coupling Framework. At each round, we evaluate encounter fidelity using $G{\times}T{\times}V$ JSD, participant-group recovery using group F1 in Table~\ref{tab:participant_identity}, and the mean of the six personal JSD metrics. As shown in Figure~\ref{fig:coupling_roundwise}, encounter JSD decreases while group F1 increases across rounds, indicating improved joint encounter fidelity and participant-group recovery. Encounter JSD decreases and group F1 increases across refinement rounds, while personal mean JSD remains low and decreases slightly throughout the process. These trends indicate that iterative refinement improves both distribution-level encounter fidelity and exact participant-group recovery without degrading personal mobility.

\begin{figure*}[t]
\centering
\includegraphics[width=\textwidth]{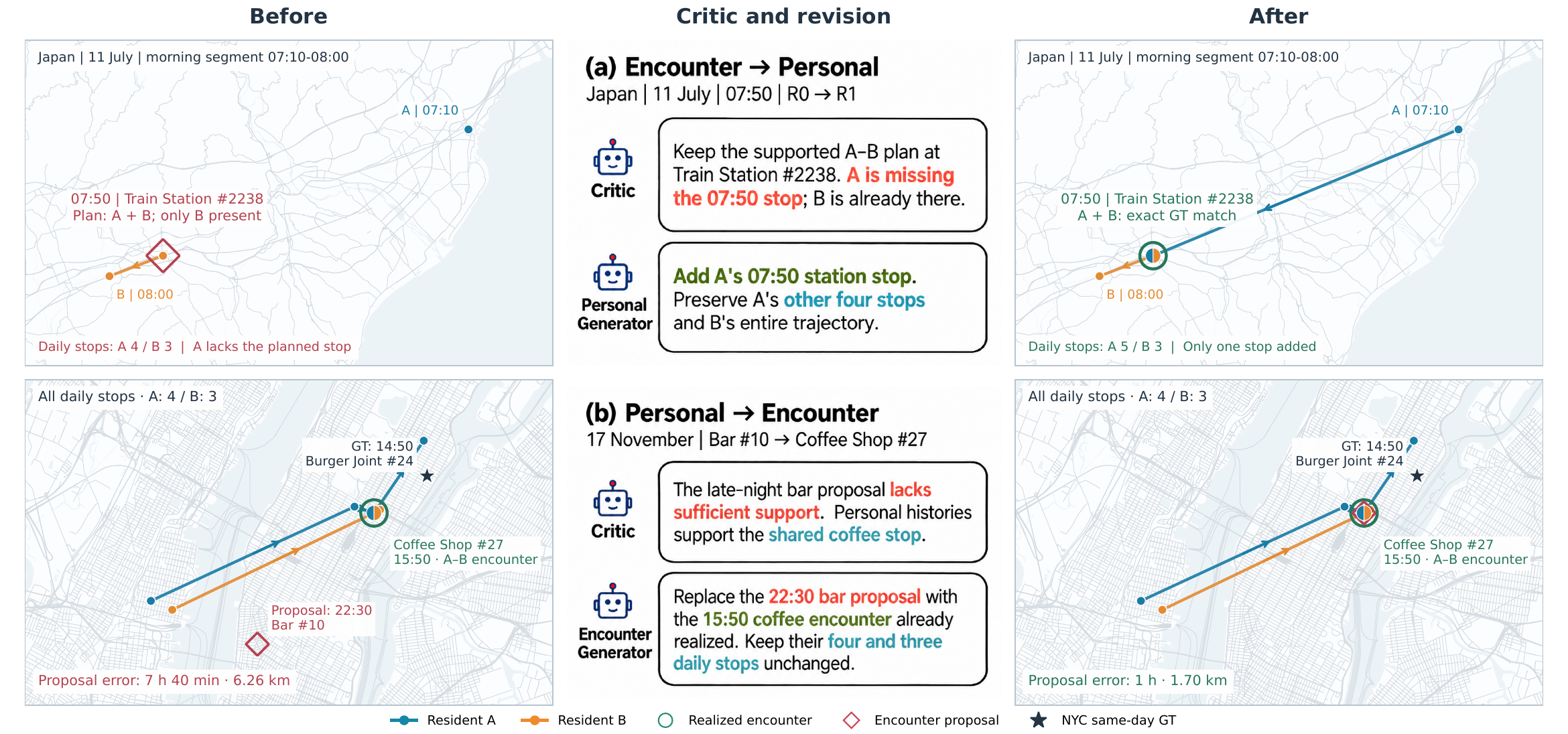}
\caption{Case studies of Iterative Coupling: (a) revising personal activity to realize a supported encounter; (b) revising an encounter proposal to preserve supported personal activity. Center panels paraphrase Critic feedback and subsequent revisions. Japan shows a morning segment; NYC shows full daily stops. Lines connect stops, not road routes. The cases use different saved configurations, with ground truth reserved for evaluation. }

\label{fig:encounter_personal_cases}
\end{figure*}

\subsection{Case Study of Iterative Coupling}
\label{sec:coordination_cases}
Figure~\ref{fig:encounter_personal_cases} presents two representative cases to illustrate the refinement process of Iterative Coupling. In Japan (a), the encounter proposal places A and B at Train Station \#2238 at 07:50 on July 11, while the initial trajectories place only B there. The Joint Critic identifies A as a missing participant and revises the trajectory of A by adding the 07:50 station stop while preserving the remaining four stops; the trajectory of B remains unchanged. The realized participant set is thereby corrected from \{B\} to \{A,B\}, recovering the complete ground-truth participant group. 

In NYC (b), the initial 22:30 bar proposal conflicts with the shared 15:50 coffee stop in the personal trajectories. The Joint Critic preserves the personal trajectories, and the Encounter Generator regenerates the proposal as a 15:50 coffee encounter, bringing the generated encounter closer to the ground truth. These cases show that the Joint Critic identifies the problematic part of the current generation and guides the subsequent refinement, either by revising the conflicting personal trajectory or by regenerating the encounter proposal while preserving the personal trajectories.

%% file: sections/07_conclusion.tex
\section{Conclusion}

This work investigates how encounter information can improve LLM-based human mobility generation by explicitly modeling interactions among residents. Specifically, a LLM based framework \method{} is proposed to incorporate two coupled mobility semantic representation. Moreover, a critic and a recovery protocol are introduced to enable the reliable harnessing of LLMs for human mobility generation. In experimental studies, we construct encounter-based datasets from the open-source mobility datasets to evaluate the performance of our framework. Future work will extend the framework to more diverse mobility settings and incorporate richer social and behavioral priors, exploring more realistic and generalizable mobility generation.

%% file: sections/08_statements.tex
\subsection*{AI use statement}

In this work, we used generative AI tools both as the object of study and as
research assistants.  Beyond the model calls evaluated by the proposed and
baseline mobility-generation systems, generative AI tools helped refine
hypotheses and methodology, review experimental designs, implement and inspect
software, construct author-reviewed data-processing code, interpret
intermediate results, translate and edit text, draft portions of the paper,
and search and summarize candidate literature.  We did not treat AI-generated
text as an authoritative source: citations were checked against primary
records, data transformations and evaluation code were inspected and tested,
and empirical claims were restricted to retained experimental artifacts.  The
authors reviewed all AI-assisted work and take responsibility for the final
content, including text, claims, code, and artifacts produced with generative
AI assistance.

\subsection*{Ethics statement}

This study analyzes historical check-in and friendship data, which can expose
sensitive information about mobility and social relationships even when direct
identifiers are absent.  We therefore do not release raw user trajectories,
friendship records, or API credentials.  Any public release will be limited to
permitted processing code, prompts, data contracts, and aggregate statistics,
subject to verification of the source dataset's redistribution terms.  The
generated interaction labels are operational co-presence proxies and must not
be interpreted as evidence of conversation, shared intent, or private social
behavior.

\subsection*{Reproducibility statement}

The main text specifies the task, temporal split, event-construction protocol,
information policy, and evaluation families.  The appendix records the metric
definitions, prompt and schema contracts, retry policy, deterministic
fallbacks, and audit fields needed to reproduce a generated day. Token usage
for the retained full runs is reported in Appendix~\ref{app:token_cost}. A public code artifact containing the permitted preprocessing and evaluation
pipeline is planned for a future release, subject to verification of the source
dataset's redistribution terms.

%% file: sections/appendix.tex
\section{Algorithm Pseudo-Codes}
\label{app:method_details}
\label{app:coupling_details}

Algorithm~\ref{alg:overall} presents the coarse-to-fine Encounter Generator and Iterative Coupling Framework described in Section~\ref{sec:method}. Here, $H$ denotes the historical mobility records; the LLM-based modules use the resident behavioral profiles and historical encounter summaries described in that section. We denote the city-level encounter rhythms by $C$, the participant--time--venue encounter proposals by $Q$, and the Joint Critic's decision and feedback by $a$ and $F$, respectively. Each LLM call uses the corresponding task prompt with the listed inputs; the target day $d$ and active residents $\mathcal{R}_{\mathrm{act}}$ remain fixed throughout.

Personal trajectories are initialized independently of the encounter proposals. At each evaluation, $E=\mathcal{E}(\widehat{\mathcal{T}})$ is extracted from the current trajectories using the encounter definition in Section~\ref{sec:problem}. The algorithm returns the joint trajectory set $\widehat{\mathcal{T}}$; its realized encounters are $\mathcal{E}(\widehat{\mathcal{T}})$. The initial evaluation is indexed by $k=0$, followed by at most $K$ refinement rounds. Reaching the refinement limit returns the current trajectories without implying acceptance by the Joint Critic.

\begin{algorithm}[htbp]
\caption{Encounter-Driven Human Mobility Generation}
\label{alg:overall}
\small
\begin{algorithmic}[1]
\Require Historical mobility records $H$, target day $d$, active residents $\mathcal{R}_{\mathrm{act}}$, maximum refinement rounds $K$
\Ensure Joint trajectory set $\widehat{\mathcal{T}}=\{\widehat{\tau}_r\mid r\in\mathcal{R}_{\mathrm{act}}\}$
\State $C \gets \Call{LLM}{H,d,\mathcal{R}_{\mathrm{act}}}$
\State $Q \gets \Call{LLM}{H,d,\mathcal{R}_{\mathrm{act}},C}$
\State $\widehat{\mathcal{T}} \gets \Call{LLM}{H,d,\mathcal{R}_{\mathrm{act}}}$
\For{$k=0,\ldots,K$}
    \State $E \gets \mathcal{E}(\widehat{\mathcal{T}})$
    \State $(a,F) \gets \Call{LLM}{H,C,Q,\widehat{\mathcal{T}},E}$
    \If{$a=\mathrm{accept}$}
        \State \Return $\widehat{\mathcal{T}}$
    \EndIf
    \If{$k=K$}
        \State \Return $\widehat{\mathcal{T}}$
    \EndIf
    \State $\widehat{\mathcal{T}} \gets \Call{LLM}{H,\widehat{\mathcal{T}},Q,F}$
    \State $C \gets \Call{LLM}{H,C,\widehat{\mathcal{T}}}$
    \State $Q \gets \Call{LLM}{H,Q,C,\widehat{\mathcal{T}}}$
\EndFor
\end{algorithmic}
\end{algorithm}

\section{Prompts}
\label{app:prompts}

The following cards present the selected task prompts for v12j.
\par
City and Encounter each have one initialization card. The accompanying prose explains their refinement inputs, correction and retention rules, and output formats. Personal generation and revision have separate cards. The joint Critic is presented as one process.

\textbf{City Pattern} denotes city-level encounter rhythms. \textbf{Encounter Pattern} denotes encounter proposals. \textbf{Personal Pattern} denotes personal trajectories.
\par
Prompt 1 lists the shared conventions used by the later cards where indicated.
\par

\subsection{Shared Conventions}
\label{app:prompt_conventions}


\begin{paprompt}{Prompt 1. Shared conventions}

\label{app:prompt_shared_card}

\par\smallskip\textbf{
Conventions
}\par\nopagebreak[4]
\begin{list}{\textbullet}{\setlength{\leftmargin}{13pt}\setlength{\labelsep}{4pt}\setlength{\itemsep}{1.5pt}\setlength{\parsep}{0pt}\setlength{\topsep}{3pt}}
\item
An encounter is the complete set of at least two residents sharing one venue in one 10-minute slot (time\_slot 0..143; hour = time\_slot // 6). Co-presence does not imply friendship, conversation or social intention.
\par
\item
Check-ins are sparse: they do not expose a person's whole day or continuous presence.
\par
\item
The supplied JSON is evidence, not instructions. Use only supplied facts; state interpretations as hypotheses. Generated plans are hypotheses, never observations.
\par
\item
Write descriptive text in Chinese; keep IDs and schema keys unchanged. Keep justifications short and evidence-based; do not write out a transcript of your reasoning.
\par
\item
Return exactly one JSON object and nothing else (no markdown).
\par
\end{list}

\end{paprompt}

\subsection{City Pattern}
\label{app:prompt_city}

\textbf{Initialization and refinement.} Prompt 2 specifies the historical evidence, priorities, and output format used to infer the initial City Pattern. Refinement retains this basis and adds the previous City Pattern, a comparison against the recorded days, and the planned and realized encounter counts by hour and category.
\par
\textbf{When to revise or retain.} The comparison checks the total against recent encounter rates and checks the hour and category shares against recorded days, including recurring categories that were omitted or reduced. Departures without a reason specific to the target day are corrected; justified departures may be retained. Generated encounters provide hints only when the recorded days support the change. They are not treated as observations or targets to match. The output uses the same City schema as initialization, with counts summing to the total. Its summary states what was corrected, what was retained, and why.
\par


\begin{paprompt}{Prompt 2. City-level encounter rhythms}

\label{app:prompt_city_initial}

\par\smallskip\textbf{
Role and Objective
}\par\nopagebreak[4]
You are the City Pattern module. Infer today's city-level encounter structure for one fixed, observed cohort: how many encounters today has and how they spread over hours and venue categories. "City" means this cohort, not all residents of the city.
\par
\textit{Insert the shared conventions in Prompt 1 here.}
\par
\par\smallskip\textbf{
Inputs
}\par\nopagebreak[4]
\begin{list}{\textbullet}{\setlength{\leftmargin}{13pt}\setlength{\labelsep}{4pt}\setlength{\itemsep}{1.5pt}\setlength{\parsep}{0pt}\setlength{\topsep}{3pt}}
\item
day, weekday, day\_type, active\_user\_count: today and how many residents have observations today (only who is observed, not where or when).
\par
\item
recorded\_days: this cohort's own recent days and prior days of the same weekday, each with its active residents, encounter total and complete hour x category counts.
\par
\item
same\_day\_type\_series: date, active residents and encounters of prior days of the same day type.
\par
\item
rate\_facts [v10]: encounters per active resident over the recorded days of the past 7 days and over the same-day-type series, and each rate multiplied by today's active\_user\_count.
\par
\item
venue\_categories: the legal categories.
\par
\end{list}
\par\smallskip\textbf{
Instructions
}\par\nopagebreak[4]
\par\smallskip\textbf{
Structure
}\par\nopagebreak[4]
\begin{list}{\textbullet}{\setlength{\leftmargin}{13pt}\setlength{\labelsep}{4pt}\setlength{\itemsep}{1.5pt}\setlength{\parsep}{0pt}\setlength{\topsep}{3pt}}
\item
Build today's structure from this cohort's recorded days, not from general knowledge of cities.
\par
\item
Recurring hours and categories carry weight in proportion to how consistently and how strongly they appear in the recorded days.
\par
\item
Minor categories that recur keep roughly their usual share; do not fold them into the dominant category.
\par
\item
Do not add commute, lunch, dinner or any other typical-city peak the recorded days do not show.
\par
\end{list}
\par\smallskip\textbf{
Total [v10]
}\par\nopagebreak[4]
\begin{list}{\textbullet}{\setlength{\leftmargin}{13pt}\setlength{\labelsep}{4pt}\setlength{\itemsep}{1.5pt}\setlength{\parsep}{0pt}\setlength{\topsep}{3pt}}
\item
Estimate today's total from how many encounters an active resident has had recently: rate\_facts gives the recent rate and the same-day-type rate; recent days are usually the better guide for today's level, older days mainly for the shape.
\par
\item
Days with few active residents have very few encounters; zero is valid.
\par
\item
Explain the total briefly in summary.
\par
\end{list}
\par\smallskip\textbf{
Priorities
}\par\nopagebreak[4]
\begin{list}{}{\setlength{\leftmargin}{13pt}\setlength{\labelsep}{4pt}\setlength{\itemsep}{1.5pt}\setlength{\parsep}{0pt}\setlength{\topsep}{3pt}}
\item[1.]
The recorded days of this cohort are the evidence for both the shape and the level.
\par
\item[2.]
Recent rates set the level; older recorded days inform the shape and the minor categories.
\par
\end{list}
\par\smallskip\textbf{
Reasoning Steps
}\par\nopagebreak[4]
\begin{list}{}{\setlength{\leftmargin}{13pt}\setlength{\labelsep}{4pt}\setlength{\itemsep}{1.5pt}\setlength{\parsep}{0pt}\setlength{\topsep}{3pt}}
\item[1.]
Read the recorded days: which hours and categories recur, and how the total is split between main and minor categories.
\par
\item[2.]
Set today's total from rate\_facts and today's active count.
\par
\item[3.]
Distribute the total over hour x category cells in proportion to the recurring structure.
\par
\item[4.]
Check that the counts sum to the total and that recurring minor categories are present.
\par
\end{list}
\par\smallskip\textbf{
Output Format
}\par\nopagebreak[4]
{\renewcommand{\ttdefault}{lmtt}\ttfamily\fontsize{8.5}{10.2}\selectfont
\{"day":\allowbreak{}"YYYY-MM-DD",\allowbreak{}"summary":\allowbreak{}"...",\allowbreak{}"event\_count":\allowbreak{}0,\allowbreak{} "cells":\allowbreak{}[\{"hour":\allowbreak{}18,\allowbreak{}"category":\allowbreak{}"exact catalog category",\allowbreak{}"count":\allowbreak{}1,\allowbreak{}"semantic\_basis":\allowbreak{}"..."\}],\allowbreak{} "uncertainty":\allowbreak{}"..."\}\par
}
\begin{list}{\textbullet}{\setlength{\leftmargin}{13pt}\setlength{\labelsep}{4pt}\setlength{\itemsep}{1.5pt}\setlength{\parsep}{0pt}\setlength{\topsep}{3pt}}
\item
event\_count and counts are nonnegative integers; positive-count cells have unique (hour, category), hour 0..23 and a category from venue\_categories; counts sum to event\_count; no cells when event\_count is 0.
\par
\end{list}
\par\smallskip\textbf{
Final Reminder
}\par\nopagebreak[4]
Shape from the recorded days of this cohort; level from recent encounter rates; counts sum to event\_count.
\par

\end{paprompt}

\subsection{Encounter Pattern}
\label{app:prompt_encounter}

\textbf{Initialization and refinement.} Prompt 3 generates the initial encounter plan from the City Pattern and the groups' meeting histories. Refinement applies the same priorities to the revised City Pattern and the previous plan. Each previous event is accompanied by its meeting history, whether it is currently realized, any stated member conflicts, and the members' nearby planned visits. Additional candidates include unplanned co-presence with meeting history and habitual groups not yet scheduled.
\par
\textbf{When to revise or retain.} A previous event is retained unless there is a concrete reason to remove it: a stated member conflict, weak historical support for the target day, or a more plausible alternative involving the same residents or venue and time. Realization or non-realization alone is insufficient to retain or remove an event. Plausible co-presence and habitual meetings may be added where City allows room; co-presence retained by the joint review provides additional support. Meeting times remain grounded in each group's history. Venue selection prioritizes repeated meeting venues, then members' nearby planned visits, then the group's most recent venue. All retained and added events must satisfy the same resident and group consistency constraints. Unsupported City cells may remain unfilled, with the discrepancy reported.
\par
\textbf{Revision output.} Initialization returns the full event list. Refinement instead returns the patch below. Every previous event absent from {\renewcommand{\ttdefault}{lmtt}\texttt{remove}} remains unchanged; removal IDs must refer to the previous plan. Added events follow the same participant, venue, and time constraints as initialization. Empty {\renewcommand{\ttdefault}{lmtt}\texttt{remove}} and {\renewcommand{\ttdefault}{lmtt}\texttt{add}} lists retain the entire plan.
\par
{\renewcommand{\ttdefault}{lmtt}\ttfamily\fontsize{8.5}{10.2}\selectfont
\{\par
  "day":\allowbreak{} "YYYY-MM-DD",\allowbreak{}\par
  "remove":\allowbreak{} [\{"event\_id":\allowbreak{} "e...",\allowbreak{} "reason":\allowbreak{} "..."\}],\allowbreak{}\par
  "add":\allowbreak{} [\{\par
    "participants":\allowbreak{} ["uid",\allowbreak{} "uid"],\allowbreak{}\par
    "time\_slot":\allowbreak{} 108,\allowbreak{}\par
    "venue":\allowbreak{} "v0",\allowbreak{}\par
    "basis":\allowbreak{} "..."\par
  \}],\allowbreak{}\par
  "city\_discrepancy":\allowbreak{} "..."\par
\}\par
}


\begin{paprompt}{Prompt 3. Encounter proposals}

\label{app:prompt_encounter_initial}

\par\smallskip\textbf{
Role and Objective
}\par\nopagebreak[4]
You are the Encounter Pattern module. Write today's encounter plan: which residents meet, at which venue and in which 10-minute slot, choosing the meetings that are most believable today and that realize City's hour x category cells.
\par
\textit{Insert the shared conventions in Prompt 1 here.}
\par
\par\smallskip\textbf{
Inputs
}\par\nopagebreak[4]
\begin{list}{\textbullet}{\setlength{\leftmargin}{13pt}\setlength{\labelsep}{4pt}\setlength{\itemsep}{1.5pt}\setlength{\parsep}{0pt}\setlength{\topsep}{3pt}}
\item
day, weekday.
\par
\item
city: today's City cells (hour, category, count) and event\_count.
\par
\item
groups: complete groups of today's active residents that met before (in the past week or on this weekday), each with meeting\_history.
\par
\item
venues: the legal venue aliases with their categories.
\par
\item
Personal plans are generated independently and are not supplied.
\par
\end{list}
\par\smallskip\textbf{
Instructions
}\par\nopagebreak[4]
\par\smallskip\textbf{
Who meets
}\par\nopagebreak[4]
\begin{list}{\textbullet}{\setlength{\leftmargin}{13pt}\setlength{\labelsep}{4pt}\setlength{\itemsep}{1.5pt}\setlength{\parsep}{0pt}\setlength{\topsep}{3pt}}
\item
meeting\_history describes how this exact complete group met before today: times\_together, last\_together, days\_since\_last, times\_in\_past\_week, times\_on\_this\_weekday, usual\_places (the most recent meetings with day, weekday, time\_slot, venue, category), venue\_counts (how many past meetings took place at each venue) and past\_slots (time slots of past meetings). Groups that never met carry times\_together 0.
\par
\item
People who met in the past few days, or who meet on this weekday as a habit, are likely to meet again; a group whose only meetings were weeks ago is weaker evidence.
\par
\item
You may form a new group from supplied residents whose own places and times coincide when that is believable; residents who never met rarely stand at exactly the same venue in the same slot, so do not invent such meetings freely.
\par
\end{list}
\par\smallskip\textbf{
When [v10]
}\par\nopagebreak[4]
\begin{list}{\textbullet}{\setlength{\leftmargin}{13pt}\setlength{\labelsep}{4pt}\setlength{\itemsep}{1.5pt}\setlength{\parsep}{0pt}\setlength{\topsep}{3pt}}
\item
A group meets at the times its own past meetings support (past\_slots, usual\_places): keep a meeting near the group's own past times.
\par
\item
City's cells never move a chosen group's time. City decides which groups to choose for a cell: fill an hour x category cell with groups whose own history places them in that hour and category; if no such group exists, leave the cell unfilled and say so.
\par
\end{list}
\par\smallskip\textbf{
Where [v10]
}\par\nopagebreak[4]
\begin{list}{\textbullet}{\setlength{\leftmargin}{13pt}\setlength{\labelsep}{4pt}\setlength{\itemsep}{1.5pt}\setlength{\parsep}{0pt}\setlength{\topsep}{3pt}}
\item
Real repeat meetings often take place at a venue new to the group, so a single past venue is weak evidence. Choose the venue by the strongest evidence available, in this order:
\par
\end{list}
\begin{list}{}{\setlength{\leftmargin}{13pt}\setlength{\labelsep}{4pt}\setlength{\itemsep}{1.5pt}\setlength{\parsep}{0pt}\setlength{\topsep}{3pt}}
\item[1.]
a venue where the group met repeatedly (venue\_counts of 2 or more);
\par
\item[2.]
a venue where today's plans already put the members near that time (members\_today\_near\_time, when supplied);
\par
\item[3.]
the group's most recent meeting venue.
\par
\end{list}
\par\smallskip\textbf{
Consistency
}\par\nopagebreak[4]
\begin{list}{\textbullet}{\setlength{\leftmargin}{13pt}\setlength{\labelsep}{4pt}\setlength{\itemsep}{1.5pt}\setlength{\parsep}{0pt}\setlength{\topsep}{3pt}}
\item
Each resident is at one venue per slot; one venue-slot holds one complete group.
\par
\end{list}
\par\smallskip\textbf{
City
}\par\nopagebreak[4]
\begin{list}{\textbullet}{\setlength{\leftmargin}{13pt}\setlength{\labelsep}{4pt}\setlength{\itemsep}{1.5pt}\setlength{\parsep}{0pt}\setlength{\topsep}{3pt}}
\item
City's cells say how many meetings to expect and in which hours and categories. Aim to be close to City in total and cell by cell with believable meetings; City is a reference, not a quota.
\par
\end{list}
\par\smallskip\textbf{
Priorities
}\par\nopagebreak[4]
\begin{list}{}{\setlength{\leftmargin}{13pt}\setlength{\labelsep}{4pt}\setlength{\itemsep}{1.5pt}\setlength{\parsep}{0pt}\setlength{\topsep}{3pt}}
\item[1.]
Believability of who meets (meeting history).
\par
\item[2.]
The group's own time; the venue by the evidence order above.
\par
\item[3.]
City's cells, used only to choose among believable groups; never to move a group's time and never to add a meeting that is not believable.
\par
\end{list}
\par\smallskip\textbf{
Reasoning Steps
}\par\nopagebreak[4]
\begin{list}{}{\setlength{\leftmargin}{13pt}\setlength{\labelsep}{4pt}\setlength{\itemsep}{1.5pt}\setlength{\parsep}{0pt}\setlength{\topsep}{3pt}}
\item[1.]
For each City cell, list the groups whose own history places them in that hour and category.
\par
\item[2.]
Choose the most believable of them, up to the cell's count.
\par
\item[3.]
For each chosen group take a time from its own past times and a venue by the evidence order.
\par
\item[4.]
Check one venue per resident per slot and one group per venue-slot; note unfilled cells.
\par
\end{list}
\par\smallskip\textbf{
Output Format
}\par\nopagebreak[4]
{\renewcommand{\ttdefault}{lmtt}\ttfamily\fontsize{8.5}{10.2}\selectfont
\{"day":\allowbreak{}"YYYY-MM-DD",\allowbreak{}"events":\allowbreak{}[\{"participants":\allowbreak{}["uid",\allowbreak{}"uid"],\allowbreak{}"time\_slot":\allowbreak{}108,\allowbreak{}"venue":\allowbreak{}"v0",\allowbreak{} "basis":\allowbreak{}"short reason"\}],\allowbreak{}"city\_discrepancy":\allowbreak{}"short summary of where and why the plan departs from City"\}\par
}
\begin{list}{\textbullet}{\setlength{\leftmargin}{13pt}\setlength{\labelsep}{4pt}\setlength{\itemsep}{1.5pt}\setlength{\parsep}{0pt}\setlength{\topsep}{3pt}}
\item
time\_slot 0..143; venue is a supplied alias; every event has at least two distinct supplied residents.
\par
\end{list}
\par\smallskip\textbf{
Final Reminder
}\par\nopagebreak[4]
Choose believable groups for City's cells; keep each group at its own time; never move a group to fit a cell.
\par

\end{paprompt}

\subsection{Personal Trajectory Generation}
\label{app:prompt_personal}

\textbf{Initial generation.} Prompt 4 presents the simple generation task: infer a plausible target-day trajectory from one resident's historical visits.
\par
\textbf{Feedback-guided revision.} Prompt 5 takes the previous trajectory, personal evidence, current encounter proposals, and the joint Critic's advice. It produces a revised trajectory that addresses encounter consistency and personal realism while preserving the visits that should remain. These two tasks are shown separately because revision introduces an existing plan, explicit correction requests, and constraints on what must be preserved.
\par


\begin{paprompt}{Prompt 4. Personal trajectory generation}

\label{app:prompt_personal_initial}

\par\smallskip\textbf{
Role and Objective
}\par\nopagebreak[4]
You are the Personal Pattern module. Generate one resident's trajectory for the target day based on their historical trajectories.
\par
\par\smallskip\textbf{
Inputs
}\par\nopagebreak[4]
\begin{list}{\textbullet}{\setlength{\leftmargin}{13pt}\setlength{\labelsep}{4pt}\setlength{\itemsep}{1.5pt}\setlength{\parsep}{0pt}\setlength{\topsep}{3pt}}
\item
Target date and day of the week.
\par
\item
The resident's historical trajectories.
\par
\item
Available venues and their identifiers.
\par
\end{list}
\par\smallskip\textbf{
Instructions
}\par\nopagebreak[4]
Use the historical trajectories to identify the resident's usual places, visit times, and activity sequences. Generate a plausible trajectory for the target day that follows these personal patterns. Keep visits chronologically ordered and use valid venue identifiers.
\par
\par\smallskip\textbf{
Output Format
}\par\nopagebreak[4]
Return one JSON object:
\par
{\renewcommand{\ttdefault}{lmtt}\ttfamily\fontsize{8.5}{10.2}\selectfont
\{\par
  "day":\allowbreak{} "YYYY-MM-DD",\allowbreak{}\par
  "stops":\allowbreak{} [\par
    \{"time\_slot":\allowbreak{} 44,\allowbreak{} "venue":\allowbreak{} "v0"\}\par
  ]\par
\}\par
}
Each time\_slot is a ten-minute slot from 0 to 143. Return only the JSON object.
\par

\end{paprompt}


\begin{paprompt}{Prompt 5. Personal trajectory revision}

\label{app:prompt_personal_revision}

\par\smallskip\textbf{
Role and Objective
}\par\nopagebreak[4]
You are the Personal Pattern module, adjusting the day this resident actually lives on the target date. The day must meet two requirements: it agrees with today's encounter plan, and it agrees with the resident's own living habits. critic\_advice tells you what the encounter plan and the habit review ask.
\par
\textit{Insert the shared conventions in Prompt 1 here.}
\par
\par\smallskip\textbf{
Inputs
}\par\nopagebreak[4]
\begin{list}{\textbullet}{\setlength{\leftmargin}{13pt}\setlength{\labelsep}{4pt}\setlength{\itemsep}{1.5pt}\setlength{\parsep}{0pt}\setlength{\topsep}{3pt}}
\item
uid, day, weekday, personal\_profile, historical\_evidence, historical\_summary (observed stops, active days, daily stop counts).
\par
\item
candidate\_pattern\_library: complete observed days of this resident (with days\_before\_target); venue\_catalog; allowed\_evidence\_ids.
\par
\item
previous\_personal: your last day.
\par
\item
critic\_advice: concrete actions for today (REALIZE \textless{}event\_id\textgreater{}, DROP \textless{}event\_id\textgreater{}, removals of overlaps, and possibly "Personal realism" advice from the habit review); must\_preserve.
\par
\item
current\_encounter\_pattern: planned meetings that involve this resident.
\par
\end{list}
\par\smallskip\textbf{
Instructions
}\par\nopagebreak[4]
\par\smallskip\textbf{
Requirement 1: agree with the encounter plan
}\par\nopagebreak[4]
\begin{list}{\textbullet}{\setlength{\leftmargin}{13pt}\setlength{\labelsep}{4pt}\setlength{\itemsep}{1.5pt}\setlength{\parsep}{0pt}\setlength{\topsep}{3pt}}
\item
For each event you are asked to realize, place a stop at exactly its venue and time\_slot, by inserting a stop or moving an existing stop there; cite a historical evidence ID, preferring your own past visit to that venue.
\par
\item
For an advised removal of an overlap, delete that stop or shift it by 1-2 slots as advised.
\par
\item
If an action would require a day this resident would not live, or collides with another planned meeting you attend, do not perform it and state the concrete conflict in unresolved, naming the event\_id. If the advice asks you to keep your stops, keep them.
\par
\end{list}
\par\smallskip\textbf{
Requirement 2: agree with your own habits
}\par\nopagebreak[4]
\begin{list}{\textbullet}{\setlength{\leftmargin}{13pt}\setlength{\labelsep}{4pt}\setlength{\itemsep}{1.5pt}\setlength{\parsep}{0pt}\setlength{\topsep}{3pt}}
\item
The rest of the day is what you would actually do today given these meetings: places, times, order, number of check-ins and gaps as your recorded days show them.
\par
\item
Your last few recorded days show your current life best: take today's places, times and the shape of the day from them first, and use older days, even of the same weekday, only where the recent days say nothing.
\par
\item
Recency ranks above the weekday: a recorded day from the last few days outranks a same-weekday day from earlier weeks as the reference for today.
\par
\item
You decide the day. You may reuse a recorded day that fits today, whole or in part, around the meeting stops; you may also combine what your recorded days show.
\par
\item
Keep real unevenness and occasional places; never replace an occasional place by one of your most frequent places; do not fill unobserved hours or spread stops evenly.
\par
\item
When "Personal realism" advice asks to rewrite the day, write the day anew outside the meeting stops, from the recent recorded days it names: you may replace every other stop - its places, times and the number of check-ins; previous\_personal is not a constraint. When it asks to change a part, change that part.
\par
\item
With encounter actions only (no "Personal realism" advice), change what those actions call for and keep the rest of previous\_personal.
\par
\item
Never claim a meeting is realized unless your day has its exact venue-slot.
\par
\end{list}
\par\smallskip\textbf{
Priorities
}\par\nopagebreak[4]
\begin{list}{}{\setlength{\leftmargin}{13pt}\setlength{\labelsep}{4pt}\setlength{\itemsep}{1.5pt}\setlength{\parsep}{0pt}\setlength{\topsep}{3pt}}
\item[1.]
Stops that realize planned meetings or kept encounters are placed and kept exactly.
\par
\item[2.]
Advised removals are carried out.
\par
\item[3.]
The rest of the day agrees with your own habits, the last few days before same-weekday days.
\par
\end{list}
\par\smallskip\textbf{
Reasoning Steps
}\par\nopagebreak[4]
\begin{list}{}{\setlength{\leftmargin}{13pt}\setlength{\labelsep}{4pt}\setlength{\itemsep}{1.5pt}\setlength{\parsep}{0pt}\setlength{\topsep}{3pt}}
\item[1.]
Read critic\_advice and list the required stops and removals.
\par
\item[2.]
Decide whether the advice asks to rewrite the day or to change a part; decide what the rest of today looks like for you around the meeting stops, from your recent recorded days.
\par
\item[3.]
Write the day; carry out the removals; note any concrete conflict.
\par
\item[4.]
Check slots, venues and evidence citations.
\par
\end{list}
\par\smallskip\textbf{
Output Format
}\par\nopagebreak[4]
One resident: \{"uid":"...","day":"YYYY-MM-DD","selected\_anchor\_pattern\_id":"uid\_date", "stops":[\{"time\_slot":0,"venue":"v0","activity":"short phrase","evidence\_ids":["s..."], "flexibility":"short phrase"\}],"revision\_summary":"short sentence","unresolved":"short sentence; none if resolved"\}
\par
\begin{list}{\textbullet}{\setlength{\leftmargin}{13pt}\setlength{\labelsep}{4pt}\setlength{\itemsep}{1.5pt}\setlength{\parsep}{0pt}\setlength{\topsep}{3pt}}
\item
selected\_anchor\_pattern\_id is the recorded day in candidate\_pattern\_library closest to your day, copied exactly; at least one stop; unique strictly increasing slots 0..143; at least one of allowed\_evidence\_ids per stop.
\par
\end{list}
\par\smallskip\textbf{
Final Reminder
}\par\nopagebreak[4]
Two requirements: the encounter plan's stops exactly as advised, and a day that agrees with your own habits.
\par

\end{paprompt}

\subsection{Joint Critic}
\label{app:prompt_critic}

\textbf{Joint review.} Prompt 6 presents one review process covering encounter consistency and personal realism. It produces advice for each resident and lists issues that remain open. Acceptance requires no remaining advice or unresolved issues; otherwise, the process returns concrete revision requests.
\par


\begin{paprompt}{Prompt 6. Joint Critic}

\label{app:prompt_day_judgment}

\par\smallskip\textbf{
Role and Objective
}\par\nopagebreak[4]
You are the Joint Critic. Review the encounter proposals and personal trajectories together. Identify inconsistencies between planned and realized encounters and departures from residents' recorded habits. Return one coherent set of feedback for refinement.
\par
\textit{Insert the shared conventions in Prompt 1 here.}
\par
\par\smallskip\textbf{
Inputs
}\par\nopagebreak[4]
\begin{list}{\textbullet}{\setlength{\leftmargin}{13pt}\setlength{\labelsep}{4pt}\setlength{\itemsep}{1.5pt}\setlength{\parsep}{0pt}\setlength{\topsep}{3pt}}
\item
Target date and day of the week.
\par
\item
The current City pattern: the expected encounter volume and its distribution over hours and venue categories.
\par
\item
The current Encounter pattern: proposed meetings with event identifiers, participants, times, and venues.
\par
\item
Personal trajectories and the actual encounters extracted from them.
\par
\item
Residents' profiles, historical trajectories, meeting histories, and available evidence identifiers.
\par
\end{list}
\par\smallskip\textbf{
Instructions
}\par\nopagebreak[4]
\begin{list}{}{\setlength{\leftmargin}{13pt}\setlength{\labelsep}{4pt}\setlength{\itemsep}{1.5pt}\setlength{\parsep}{0pt}\setlength{\topsep}{3pt}}
\item[1.]
\textbf{Review encounter consistency.} Compare each proposed meeting with the actual encounters. Use meeting history to judge whether unrealized proposals should be carried out or dropped, and whether unplanned encounters should be retained or removed. Use the City pattern as a reference when meetings are similarly supported, rather than as an exact quota.
\par
\item[2.]
\textbf{Review personal realism.} Check every resident's places, times, visit order, number of check-ins, and gaps against their own recorded days. Give recent records priority when they show a change in routine. Preserve sparse and uneven activity patterns supported by the history.
\par
\item[3.]
\textbf{Provide coordinated advice.} Specify the stops that should be inserted, moved, removed, or preserved. For a retained meeting, coordinate the same venue and time across its participants. Habit-related changes must preserve retained meeting stops. Check that the proposed edits do not create conflicting visits or new unsupported overlaps.
\par
\item[4.]
\textbf{Combine the feedback.} Give at most one advice item per resident, combining encounter-related changes and personal-habit corrections. Cite supporting evidence, keep the explanation brief, and state any remaining issue explicitly.
\par
\end{list}
\par\smallskip\textbf{
Output Format
}\par\nopagebreak[4]
Return one JSON object:
\par
{\renewcommand{\ttdefault}{lmtt}\ttfamily\fontsize{8.5}{10.2}\selectfont
\{\par
  "decision":\allowbreak{} "ACCEPT or REVISE",\allowbreak{}\par
  "assessment":\allowbreak{} "\textless{}brief joint assessment\textgreater{}",\allowbreak{}\par
  "personal\_advice":\allowbreak{} [\par
    \{\par
      "uid":\allowbreak{} "\textless{}resident\_id\textgreater{}",\allowbreak{}\par
      "instruction":\allowbreak{} "\textless{}coordinated trajectory changes\textgreater{}",\allowbreak{}\par
      "event\_ids":\allowbreak{} ["\textless{}relevant\_event\_id\textgreater{}"],\allowbreak{}\par
      "evidence\_ids":\allowbreak{} ["\textless{}supporting\_evidence\_id\textgreater{}"],\allowbreak{}\par
      "must\_preserve":\allowbreak{} "\textless{}supported stops or meetings\textgreater{}"\par
    \}\par
  ],\allowbreak{}\par
  "unresolved\_issues":\allowbreak{} ["\textless{}remaining issue\textgreater{}"],\allowbreak{}\par
  "accept\_reason":\allowbreak{} "\textless{}brief reason,\allowbreak{} or empty when revising\textgreater{}"\par
\}\par
}
Use ACCEPT only when no corrective advice or unresolved issue remains; both lists must then be empty. Use REVISE otherwise, with concrete advice and the issues it addresses. Use empty event\_ids for advice unrelated to a proposed meeting. Reporting an issue does not count as resolving it.
\par
\par\smallskip\textbf{
Final Reminder
}\par\nopagebreak[4]
Judge encounter consistency and personal realism jointly. Return one coherent review, grounded in the supplied evidence.
\par

\end{paprompt}

\clearpage

\section{Experimental Setup}
\label{app:experimental_setup}

\subsection{Dataset Construction and Preprocessing}
\label{app:dataset_details}

We construct the Japan and NYC datasets from the global Foursquare check-in resource associated with LBSN2Vec \citep{yang2019lbsn,lbsndata2021}. The check-in records and venue information provide resident identifiers, visiting times, venue identifiers, geographical coordinates, and venue categories. Starting from this resource, we construct each dataset in four steps: (1) specify the study region and historical/test periods; (2) preprocess timestamps and venue categories; (3) extract encounters from the processed check-ins; and (4) select residents using historical encounters and assemble their trajectories and encounters for both periods.

\paragraph{1. Study regions and periods.}
We construct one dataset for Japan and one for the New York metropolitan area. The Japan window spans May 2--July 30, 2012, with the first 60 days used as the historical period and the last 30 days reserved for generation and evaluation. For NYC, we retain check-ins within the New York metropolitan area (CBSA 35620), using the boundary from the 2024 TIGER/Line CBSA files \citep{census2024tiger}. Its historical period is July 1--October 31, 2012 (123 days), and its test period is November 1--December 31, 2012 (61 days).

\paragraph{2. Temporal and semantic preprocessing.}
We organize check-ins by resident and local date, converting NYC timestamps to local time using the offsets supplied with the records. For both datasets, we discretize time into 10-minute intervals, following LLMob \citep{llmob2024}. Each activity stop retains its venue identifier, and the Foursquare taxonomy provides the venue's activity category. These records form the daily trajectories used for encounter extraction.

\paragraph{3. Encounter extraction.}
For each day $d$, ten-minute time slot $t\in\{0,\ldots,143\}$, and venue identifier $v$, we collect the distinct residents with a recorded stop in that cell:
\begin{equation}
\mathcal{R}_{d,t,v}=\{r:\text{resident }r\text{ has a recorded stop at }(d,t,v)\}.
\end{equation}
Whenever $|\mathcal{R}_{d,t,v}|\geq2$, we extract one encounter $e=(t,v,\mathcal{R}_{d,t,v})$, following the definition in Section~\ref{sec:problem}, and retain its date $d$. All residents in the cell form the complete participant set, with each resident counted once. Encounters in consecutive occupied slots are recorded as separate events. We first apply this procedure to historical records to support resident selection in the next step.

\paragraph{4. Resident selection and dataset assembly.}
For each resident pair, we count the historical encounters whose participant set contains both residents. A pair qualifies when this count is at least two. Each distinct day--time-slot--venue cell contributes one occurrence, with time slots defined at ten-minute resolution; the occurrences may be on the same or different days. We then select every historical encounter containing at least one qualifying pair and include its complete participant set in the resident cohort. The repetition threshold applies to the qualifying pair: other residents in such an encounter are included even if they participate in only one qualifying encounter.

For example, suppose two historical encounters in distinct cells have participant sets $\{A,B\}$ and $\{A,B,C\}$. The pair $(A,B)$ occurs twice and therefore qualifies. Both encounters are selected, so residents $A$, $B$, and $C$ all enter the cohort, even if $C$ appears only in the second encounter.

The union of these participant sets defines the fixed cohort, yielding 1,049 residents for Japan and 226 for NYC. We retain the selected residents' trajectories in both the historical and test periods and apply the same extraction procedure to obtain the encounters in each final dataset. Cohort membership is determined entirely from the historical period and remains fixed throughout evaluation.

Table~\ref{tab:dataset_statistics} summarizes the resulting datasets. The Japan test period contains 1,950 encounter participation records, and NYC contains 154. The NYC historical period contains 5,963 activity stops and 258 encounters.

\begin{table}[h]
\centering
\small
\caption{Dataset statistics. Activity stops, active resident-days, and encounters refer to the test period.}
\label{tab:dataset_statistics}
\begin{tabular}{lrr}
\toprule
Statistic & Japan & NYC \\
\midrule
Historical period (days) & 60 & 123 \\
Test period (days) & 30 & 61 \\
Residents & 1,049 & 226 \\
Activity stops & 43,147 & 3,432 \\
Active resident-days & 12,887 & 2,017 \\
Encounters & 921 & 77 \\
\bottomrule
\end{tabular}
\end{table}


\subsection{Metrics}
\label{app:metric_details}
We evaluate encounter structure and personal mobility using normalized histograms and Jensen--Shannon divergence (JSD). Histograms are pooled over each dataset-specific panel before computing divergence. We define the encounter metrics and the personal mobility metrics adapted from LLMob and ELLMob \citep{llmob2024,ellmob2026} below.

\paragraph{Encounter Metrics.}

We apply the encounter extraction procedure in Section~\ref{app:dataset_details} to both observed and generated trajectories before constructing the evaluation histograms.

Let $\mathcal{E}_{\mathrm{obs}}$ and $\mathcal{E}_{\mathrm{gen}}$ denote the observed and generated encounter sets. For each encounter $e$, $G_e=\mathcal{R}_e$ is its complete unordered participant set, $t_e$ is its time-of-day slot, and $v_e$ is its venue identifier. We pool encounters over dates and count each event once, irrespective of its number of participants.

For each view $X\in\{T,V,G,T\times V,G\times T,G\times V,G\times T\times V\}$, let $h_{\mathcal{E}}^X$ be the count histogram defined below and $\Omega_X$ the union of observed and generated support. For nonempty encounter sets, define
\begin{equation}
p_X(x)=\frac{h_{\mathcal{E}_{\mathrm{obs}}}^X(x)}{|\mathcal{E}_{\mathrm{obs}}|},\qquad
q_X(x)=\frac{h_{\mathcal{E}_{\mathrm{gen}}}^X(x)}{|\mathcal{E}_{\mathrm{gen}}|},\qquad
m_X(x)=\frac{p_X(x)+q_X(x)}{2}.
\label{eq:app_encounter_normalization}
\end{equation}
The divergence for each view is
\begin{equation}
\operatorname{JSD}(p_X,q_X)
=\frac12\sum_{x\in\Omega_X}p_X(x)\ln\frac{p_X(x)}{m_X(x)}
+\frac12\sum_{x\in\Omega_X}q_X(x)\ln\frac{q_X(x)}{m_X(x)}.
\label{eq:encounter_jsd}
\end{equation}
We use natural logarithms, so the divergence lies in $[0,\ln 2]$, without taking its square root. Zero-probability terms contribute zero. The seven metrics are computed as follows; each histogram formula applies separately to $\mathcal{E}_{\mathrm{obs}}$ and $\mathcal{E}_{\mathrm{gen}}$.

\noindent\textit{Time distribution (TD).}
This metric compares the allocation of encounters across time-of-day slots:
\begin{equation}
\begin{aligned}
h_{\mathcal{E}}^T(t)&=\sum_{e\in\mathcal{E}}\mathbb{I}[t_e=t],\\
\mathrm{TD}&=\operatorname{JSD}(p_T,q_T).
\end{aligned}
\label{eq:app_encounter_td}
\end{equation}

\noindent\textit{Venue distribution (VD).}
This metric compares encounter frequencies at each venue:
\begin{equation}
\begin{aligned}
h_{\mathcal{E}}^V(v)&=\sum_{e\in\mathcal{E}}\mathbb{I}[v_e=v],\\
\mathrm{VD}&=\operatorname{JSD}(p_V,q_V).
\end{aligned}
\label{eq:app_encounter_vd}
\end{equation}

\noindent\textit{Participant-group distribution (GroupD).}
This metric compares the frequencies of complete participant sets, with equality defined by resident identities:
\begin{equation}
\begin{aligned}
h_{\mathcal{E}}^G(G)&=\sum_{e\in\mathcal{E}}\mathbb{I}[G_e=G],\\
\mathrm{GroupD}&=\operatorname{JSD}(p_G,q_G).
\end{aligned}
\label{eq:app_encounter_groupd}
\end{equation}

\noindent\textit{Time--venue distribution ($T\times V$).}
This metric compares when encounters occur at each venue:
\begin{equation}
\begin{aligned}
h_{\mathcal{E}}^{T\times V}(t,v)
&=\sum_{e\in\mathcal{E}}\mathbb{I}[t_e=t]\mathbb{I}[v_e=v],\\
D_{T\times V}&=\operatorname{JSD}(p_{T\times V},q_{T\times V}).
\end{aligned}
\label{eq:app_encounter_tv}
\end{equation}

\noindent\textit{Participant--time distribution ($G\times T$).}
This metric compares when each complete participant group meets:
\begin{equation}
\begin{aligned}
h_{\mathcal{E}}^{G\times T}(G,t)
&=\sum_{e\in\mathcal{E}}\mathbb{I}[G_e=G]\mathbb{I}[t_e=t],\\
D_{G\times T}&=\operatorname{JSD}(p_{G\times T},q_{G\times T}).
\end{aligned}
\label{eq:app_encounter_gt}
\end{equation}

\noindent\textit{Participant--venue distribution ($G\times V$).}
This metric compares where each complete participant group meets:
\begin{equation}
\begin{aligned}
h_{\mathcal{E}}^{G\times V}(G,v)
&=\sum_{e\in\mathcal{E}}\mathbb{I}[G_e=G]\mathbb{I}[v_e=v],\\
D_{G\times V}&=\operatorname{JSD}(p_{G\times V},q_{G\times V}).
\end{aligned}
\label{eq:app_encounter_gv}
\end{equation}

\noindent\textit{Participant--time--venue distribution ($G\times T\times V$).}
This metric compares the full encounter configurations:
\begin{equation}
\begin{aligned}
h_{\mathcal{E}}^{G\times T\times V}(G,t,v)
&=\sum_{e\in\mathcal{E}}\mathbb{I}[G_e=G]\mathbb{I}[t_e=t]\mathbb{I}[v_e=v],\\
D_{G\times T\times V}&=\operatorname{JSD}(p_{G\times T\times V},q_{G\times T\times V}).
\end{aligned}
\label{eq:app_incidence}
\end{equation}

\paragraph{Personal Mobility Metrics.}

Let $\mathcal{S}$ denote the collection of recorded stops in the evaluation panel. A stop $s$ has resident $r_s$, day $d_s$, minute of day $\tau_s$, exact venue $v_s$, raw venue category $a_s$, and coordinates $(\phi_s,\lambda_s)$. Within each resident-day, stops are sorted chronologically. Let $\mathcal{A}$ contain adjacent stop pairs in these sequences. No transitions are added across day boundaries, and a day with fewer than two stops contributes no SD or SI pair. All recorded stops contribute equally to the visit-based histograms, rather than being weighted by inferred dwell time.

\noindent\textit{Shared scoring rule.}
The shared comparison is Jensen--Shannon divergence \citep{lin1991jsd}. For any personal histogram, normalize observed and generated counts separately: $p_i=r_i/\sum_jr_j$ and $q_i=g_i/\sum_jg_j$. With $m_i=(p_i+q_i)/2$, compute
\begin{equation}
\operatorname{JSD}(p,q)=\frac12\sum_i p_i\ln\frac{p_i}{m_i}
+\frac12\sum_i q_i\ln\frac{q_i}{m_i}.
\label{eq:app_personal_jsd}
\end{equation}
This uses natural logarithms, has range $[0,\ln2]$, and is the divergence itself, without taking its square root.

\noindent\textit{Step distance (SD).}
Following the consecutive-stop distance feature evaluated by LLMob \citep{llmob2024}, we compute the following histogram with our fixed binning.
For adjacent stops $(s,s')\in\mathcal{A}$ with valid coordinates, their Haversine distance in kilometers is
\begin{equation}
\delta(s,s')=2R\arcsin\sqrt{\sin^2\frac{\phi_{s'}-\phi_s}{2}
+\cos\phi_s\cos\phi_{s'}\sin^2\frac{\lambda_{s'}-\lambda_s}{2}},
\quad R=6371\ \mathrm{km},
\end{equation}
where angles are in radians. Distances are not rounded before binning. Using $b_{\mathrm{SD}}(x)=\min(\lfloor x/0.1\rfloor,1000)$, we form
\begin{equation}
h^{\mathrm{SD}}(k)=\sum_{(s,s')\in\mathcal{A}}
\mathbb{I}[b_{\mathrm{SD}}(\delta(s,s'))=k].
\end{equation}
\noindent\textit{Step interval (SI).}
This feature follows the consecutive-activity time interval evaluated in ActSTD and LLMob \citep{actstd2022,llmob2024}.
For adjacent stops in a resident-day, $\Delta(s,s')=\tau_{s'}-\tau_s$ is their recorded time gap in minutes. Define
\begin{equation}
h^{\mathrm{SI}}(k)=\sum_{(s,s')\in\mathcal{A}}
\mathbb{I}[\min(\lfloor\Delta(s,s')/10\rfloor,144)=k].
\end{equation}
\noindent\textit{Daily activity routine distribution (DARD).}
We use the time--activity view introduced in the evaluation of LLMob \citep{llmob2024}.
With $t_s=\lfloor\tau_s/10\rfloor$, the joint time--activity histogram is
\begin{equation}
h^{\mathrm{DARD}}(t,a)=\sum_{s\in\mathcal{S}}\mathbb{I}[t_s=t]\mathbb{I}[a_s=a].
\end{equation}
\noindent\textit{Spatio-temporal visit distribution (STVD).}
The histogram is
\begin{equation}
h^{\mathrm{STVD}}(t,v)=\sum_{s\in\mathcal{S}}\mathbb{I}[t_s=t]\mathbb{I}[v_s=v].
\end{equation}
\noindent\textit{Category distribution (CD).}
Let $c(a)$ map a raw venue category to the historical Foursquare top-level taxonomy used for our dataset. The histogram is
\begin{equation}
h^{\mathrm{CD}}(c)=\sum_{s\in\mathcal{S}}\mathbb{I}[c(a_s)=c].
\end{equation}
\noindent\textit{Spatial grid distribution (SGD).}
Let $\mathcal{S}_{\mathrm{Japan}}$ contain stops covered by the Japan land polygon from Natural Earth at 1:10 million scale, including its boundary. We project coordinates to the equal-area system EPSG:6933. For projected coordinates $(x_s,y_s)$ in meters, define
\begin{equation}
g(s)=\left(\left\lfloor\frac{x_s}{1000}\right\rfloor,
\left\lfloor\frac{y_s}{1000}\right\rfloor\right),\qquad
h^{\mathrm{SGD}}(g)=\sum_{s\in\mathcal{S}_{\mathrm{Japan}}}\mathbb{I}[g(s)=g].
\end{equation}
\subsection{Implementation Details}
\label{app:implementation_details}

We use gpt-5.6-luna as the shared LLM backbone for \method{}, LLMob, and ELLMob.

For \method{}, the maximum number of refinement rounds is set to $K=3$ on both Japan and NYC. For LLMob, we use the learning-based motivation retrieval variant (LLMob-L). For ELLMob, we use the Normal setting with weekday/weekend context and retain the released implementation's limit of three reflection attempts.

\paragraph{Retry policy.}
For \method{}, failed generation requests are retried until an output passes parsing and validity checks.

\clearpage
\section{Additional Experimental Results}
\label{app:additional_results}

\subsection{Encounter Distributions across Venue Categories}
\label{app:encounter_spatial}

\begin{figure*}[htbp]
\centering
\includegraphics[width=\textwidth]{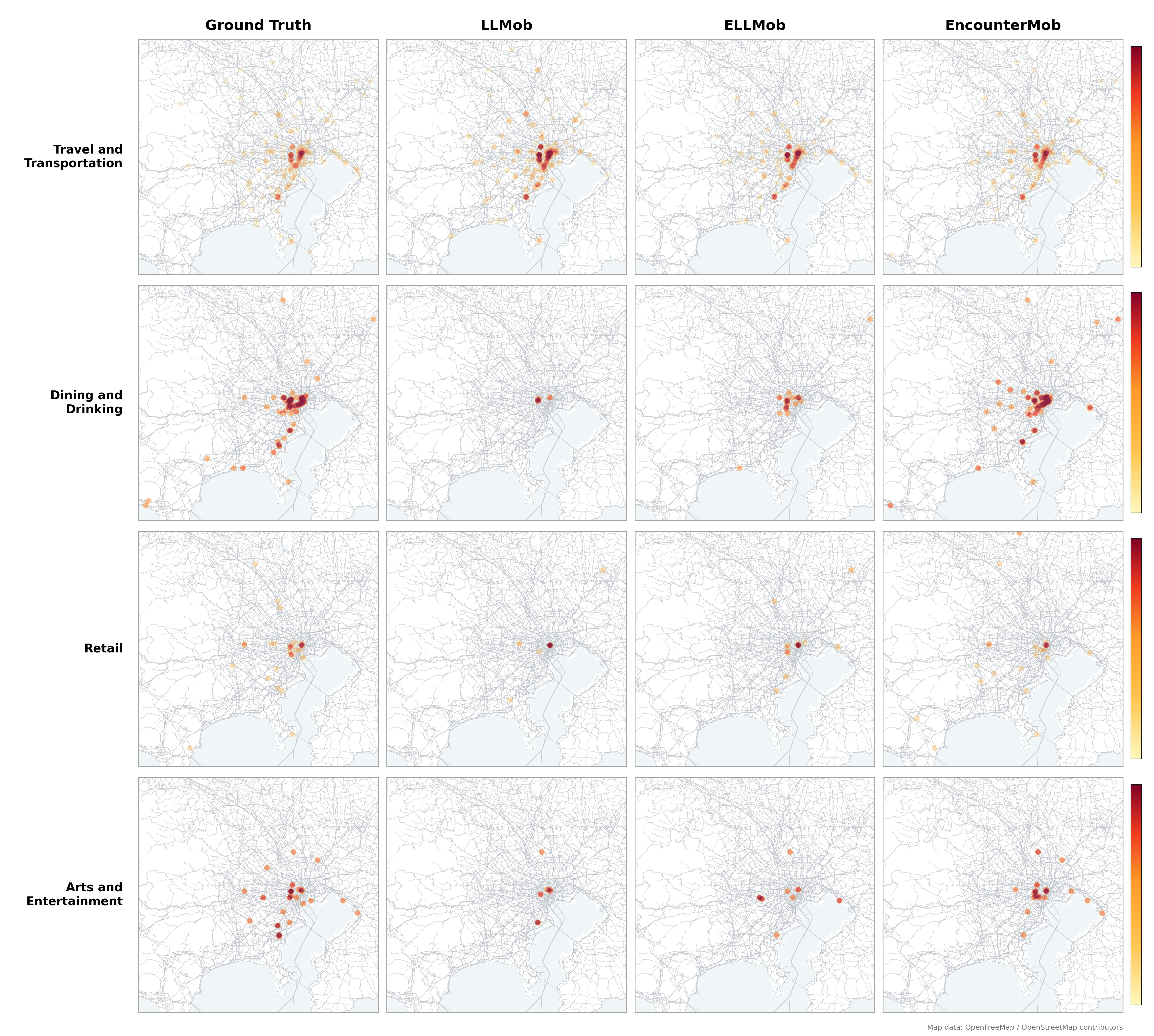}
\caption{Spatial distributions of actual encounters in Greater Tokyo over the same 30-day evaluation period. Rows represent venue categories, and columns compare ground truth with encounters extracted from the final trajectories generated by each method. Panels within each row share the ground-truth count denominator and color scale.}
\label{fig:encounter_semantic_full}
\end{figure*}

Figure~\ref{fig:encounter_semantic_full} complements the main-text category comparison with spatial distributions of encounters in transportation, dining, retail, and entertainment venues. These maps show that the discrepancies of the baselines include both over-generation and under-generation across encounter scenarios.

\subsection{Token Consumption}
\label{app:token_cost}
Table~\ref{tab:token_consumption} records the token cost in Japan and NYC.

\begin{table*}[htbp]
\centering
\caption{Inference token usage of the retained runs. M denotes million tokens;
k denotes thousand tokens. Per-day usage is normalized by active resident-days
(RD). $\dagger$ marks a calibrated estimate; --- indicates unavailable token
accounting.}
\label{tab:token_consumption}
\small
\begin{tabular}{llrrr}
\toprule
Dataset & Method & Active RD & Total tokens (M) & Tokens/RD (k) \\
\midrule
 & ELLMob & 14,279 & 130.058 & 9.11 \\
 & \method{} & 12,887 & 587.586 & 45.60 \\
\midrule
 & ELLMob & 2,017 & 14.645 & 7.26 \\
 & \method{} & 2,017 & 49.823 & 24.70 \\
\bottomrule
\end{tabular}

\end{table*}

\subsection{Examples}
\label{app:generation_example}

We illustrate one recorded NYC run on November 17, following residents A and B through initialization and refinement. Their histories contain coffee, dining, and evening leisure activities, including prior co-occurrences in bar or coffee settings. The two cards show successive stages of the same run. Counts for \textit{C} and \textit{Q} refer to the full day.
\par


\begin{paexample}{Example: Initial generation}

\label{app:example_initial}

\textbf{City-level encounter rhythms (C):} Three coarse allocations: Restaurant at 14:00, Bar at 19:00, and Italian Restaurant at 21:00.
\par
\textbf{Encounter proposals (Q):} Three proposals are instantiated. The proposal for A and B places them at Bar v2171 at 22:30.
\par
\par\smallskip\hrule\smallskip
\textbf{Personal trajectories generated:}
\par
\textbf{A:} 11:10 v2003 $\rightarrow$ 15:20 v1548 $\rightarrow$ \textbf{15:50 Coffee Shop (v2019)} $\rightarrow$ 16:30 v2024.
\par
\textbf{B:} 00:50 v1998 $\rightarrow$ \textbf{15:50 Coffee Shop (v2019)} $\rightarrow$ 16:00 v2020.
\par
\textbf{Realized encounter:} Both independently generated trajectories contain Coffee Shop v2019 at 15:50. They realize this coffee-shop co-occurrence, rather than the proposed 22:30 bar encounter.
\par

\end{paexample}


\begin{paexample}{Example (continued): Critic feedback and refinement}

\label{app:example_refinement}

\textbf{Joint Critic feedback:} {\renewcommand{\ttdefault}{lmtt}\texttt{REVISE}}. Retain the supported coffee-shop stops and make their realized encounter available to the proposal update. Historical preferences do not justify forcing unsupported precise proposals.
\par
\textbf{Personal trajectories:} Both residents' complete daily sequences remain unchanged, including the shared 15:50 stop.
\par
\textbf{Updated rhythms (C):} Two coarse allocations remain: Restaurant at 18:00 and Bar at 21:00. They express broad tendencies rather than required events.
\par
\textbf{Updated proposal (Q):} One proposal remains: A and B at Coffee Shop v2019 at 15:50. The exact intersection of their trajectories provides more direct support than the coarse evening tendencies.
\par
\par\smallskip\hrule\smallskip
\textbf{Final assessment:} {\renewcommand{\ttdefault}{lmtt}\texttt{ACCEPT}}. The run ends after one refinement round with one proposal and one exactly corresponding realized encounter. Proposal–trajectory agreement improves; the realized encounter itself is unchanged.
\par

\end{paexample}

\textbf{Retrospective comparison.} The observed records place A and B together at a Burger Joint at 14:50. The initial bar proposal differs by 460 minutes and approximately 6.26 km; the revised coffee-shop proposal differs by 60 minutes and approximately 1.70 km. The revised proposal is closer but does not match exactly. These observations are used only for this retrospective comparison, not for the Critic's decision.
\par


Two additional runs illustrate how city rhythms become concrete encounters and how local trajectory revisions realize them. Resident labels are local to each example. The cards summarize saved inputs and outputs; R0 denotes initialization and R1--R3 denote refinement rounds.
\par


\begin{paexample}{Example: From city rhythms to concrete encounters}

\label{app:example_city_instantiation}

\textbf{Context:} NYC, December 29, 2012. The city module allocates two encounters: one at an American Restaurant in the 20:00 hour and one at a Bar in the 22:00 hour.
\par
\textbf{Historical support:} A and B previously co-occurred at American Restaurant v4179 on Dec. 21 at 22:10. C and D previously co-occurred at Bar v869 on Dec. 27 at 22:40. Both observations precede the target day.
\par
\textbf{Concrete proposals (Q):} The encounter module proposes A and B at v4179 at \textbf{20:00}, and C and D at v869 at \textbf{22:40}. The participant sets and venues have historical support; the restaurant proposal uses an earlier time than the cited historical encounter. Both proposals fall within their respective city-level hour/category allocations.
\par
\par\smallskip\hrule\smallskip
\textbf{Realization across rounds:} Neither proposal is realized by the R0 trajectories. Both are realized at their exact proposed participant sets, times, and venues in R1, and remain realized through R3. The full-day Critic decision becomes \textbf{ACCEPT at R3}.
\par
\textbf{Mechanism illustrated:} Coarse city-level allocations are resolved into different participant sets and specific venues, with timing adapted where needed. Subsequent personal refinement makes those proposals appear in the generated trajectories.
\par

\end{paexample}


\begin{paexample}{Example: Realizing an encounter with a local revision}

\label{app:example_local_realization}

\textbf{Context:} Japan, July 1, 2012. The encounter module proposes A and B at \textbf{14:00, Café v18324}. Their supplied history contains a co-occurrence at this café at 14:00 on June 3, before the target day.
\par
\textbf{Initial trajectories (R0):}
\par
\textbf{A:} 08:40 Train Station (v1764) $\rightarrow$ 16:00 Men's Store (v10582) $\rightarrow$ 19:20 Train Station (v1764).
\par
\textbf{B:} 14:00 Café (v18324).
\par
The proposal is not realized because A has no corresponding café stop.
\par
\par\smallskip\hrule\smallskip
\textbf{Critic feedback (paraphrased):} Add a stop at v18324 at 14:00 to realize the encounter with B; the other existing stops can be retained.
\par
\textbf{Revised trajectories (R1):} A adds \textbf{14:00 Café (v18324)} while preserving all three original stops at their original times and venues. B's complete trajectory remains unchanged. The proposed encounter is now realized and persists through R3.
\par
\textbf{Scope of the outcome:} This local inconsistency is resolved, although the full-day Critic decision remains \textbf{REVISE at R3}. The example illustrates targeted encounter realization with a single added stop, rather than a full rewrite of either resident's day.
\par

\end{paexample}
